%% file: IMAE_iclr2023_workshop.tex
\documentclass{article} % For LaTeX2e
\usepackage{iclr2023_conference,times}

\input{math_commands.tex}

\usepackage{hyperref}
\usepackage{url}

\usepackage{graphicx}
\graphicspath{{Figures/}}

\usepackage{listings}
\usepackage{xcolor,colortbl}

\definecolor{Gray}{gray}{0.85}
\definecolor{LightCyan}{rgb}{0.48,0.78,1}
\definecolor{LightColor1}{rgb}{0.68,0.88,1.2}

\definecolor{LightGray}{gray}{.8}
\newcolumntype{C}[1]{>{\centering\arraybackslash}p{#1}}

\newcolumntype{y}{>{\columncolor{yellow}}c}
\newcolumntype{a}{>{\columncolor{lightgray}}c}
\newcolumntype{g}{>{\columncolor{pink}}c}
\newcolumntype{o}{>{\columncolor{LightColor1}}c}
\newcolumntype{u}{>{\columncolor{LightCyan}}c}

\newcommand{\specialcell}[2][c]{%
	\begin{tabular}[#1]{@{}c@{}}#2\end{tabular}}

\PassOptionsToPackage{hyphens}{url}
\usepackage[T1]{fontenc}

\usepackage{amsmath,amssymb,amsfonts} 
\usepackage{fixltx2e}
\usepackage{multirow}
\usepackage{booktabs}
\usepackage{amsmath}

\usepackage{subcaption}

\usepackage{makecell}
\usepackage[T1]{fontenc}

\lstdefinestyle{customc}{
	belowcaptionskip=1\baselineskip,
	breaklines=true,
	frame=L,
	xleftmargin=\parindent,
	language=C,
	showstringspaces=false,
	basicstyle=\footnotesize\ttfamily,
	keywordstyle=\bfseries\color{green!40!black},
	commentstyle=\itshape\color{purple!40!black},
	identifierstyle=\color{blue},
	stringstyle=\color{orange},
}

\lstdefinestyle{customasm}{
	belowcaptionskip=1\baselineskip,
	frame=L,
	xleftmargin=\parindent,
	language=[x86masm]Assembler,
	basicstyle=\footnotesize\ttfamily,
	commentstyle=\itshape\color{purple!40!black},
}

\usepackage{soul}
\definecolor{mycolor}{rgb}{0.75, 0.75, 0.75}
\definecolor{lavendergray}{rgb}{0.77, 0.76, 0.82}
\definecolor{lightgray}{rgb}{0.83, 0.83, 0.83}

\usepackage[export]{adjustbox}
\usepackage{multicol}
\title{
\vspace{-0.2cm}
IMAE for Noise-Robust Learning: Mean Absolute Error Does Not Treat Examples Equally and Gradient Magnitude's Variance Matters
\vspace{-0.40cm}
}

\author{%
	Xinshao Wang\thanks{
		This work was mainly done at Queen's University Belfast and University of Oxford. 
	}
	~\thanks{
		For the source code, based on the requests for academic research and kindness to cite our work, we will release and maintain it in \href{https://github.com/XinshaoAmosWang/DeepCriticalLearning}{https://github.com/XinshaoAmosWang/DeepCriticalLearning}.
	} 
	\\
	University of Oxford\\ 
	\texttt{xinshaowang@gmail.com} \\
	\And
	Yang Hua \\
	Queen's University Belfast\\
	\texttt{y.hua@qub.ac.uk~~~~~~~} \\
	
	\AND
	Elyor Kodirov \\
	~\\
	\texttt{ekodirov@gmail.com } \\
	
	\And
	David A. Clifton\thanks{
		Prof. David A. Clifton was supported by the NIHR Oxford Biomedical Research Centre, the InnoHK Hong Kong Centre for Cerebro-cardiovascular Health Engineering (COCHE), and the Pandemic Sciences Institute at the University of Oxford.
		Prof. David A. Clifton was also funded by an NIHR Research Professorship and an RAEng Research Chair. 
	}\\
	University of Oxford\\
	\texttt{davidc@robots.ox.ac.uk}\\
	
	\And
	Neil M. Robertson \\
	Queen's University Belfast \\
	\texttt{n.robertson@qub.ac.uk} 
}

\iclrfinalcopy % Uncomment for camera-ready version, but NOT for submission.
\begin{document}

\maketitle

\vspace{-0.40cm}
%%%%%%%%% ABSTRACT
\begin{abstract}
	\vspace{-0.30cm}
	In this work, we study robust deep learning against abnormal training data from the perspective of example weighting built in empirical loss functions, i.e., gradient magnitude with respect to logits, an angle that is not thoroughly studied so far. 
	Consequently, we have two key findings: 
	(1) Mean Absolute Error (MAE) Does Not Treat Examples Equally. We present new observations and insightful analysis about MAE, which is theoretically proved to be noise-robust.
	First, we reveal its underfitting problem in practice.   
	Second, we analyse that MAE's noise-robustness is from emphasising on uncertain examples instead of treating training samples equally, as claimed in prior work.   
	%Third, its underfitting issue is interpreted from the differentiation degree viewpoint. 
	%We provide a theoretical and empirical analysis for a crucial research question: 
	%Theoretically, compared with categorical cross entropy (CCE), mean absolute error (MAE) is proved to more noise-tolerant. In practice, CCE can fit training data perfectly even when labels are random, thus being noise-sensitive. While MAE cannot even fit clean patterns well under severe noise.
	%What is the reason?
	%However, 
	%empirically we observe that MAE underfits to even clean labels when severe noise exists.  
	% thus being inferior to CCE especially when CCE is assisted with early stopping in practice.  
	%We study why this happens .  
	%
	%\\
	(2) The Variance of Gradient Magnitude Matters. We propose an effective and simple solution to enhance MAE's fitting ability while preserving its noise-robustness. 
	Without changing MAE's overall weighting scheme, i.e., what examples get higher weights, we simply change its weighting variance non-linearly so that the impact ratio between two examples are adjusted. 
	Our solution is termed Improved MAE (IMAE). 
	We prove IMAE's effectiveness using extensive experiments: image classification under clean labels, synthetic label noise, and real-world unknown noise.   
	%We conclude IMAE is superior to CCE, the most popular loss for training DNNs. 
	%
	%Source code: \href{https://github.com/XinshaoAmosWang/Improving-Mean-Absolute-Error-against-CCE}{https://github.com/XinshaoAmosWang/Improving-Mean-Absolute-Error-against-CCE}   
	%=> To do: example's weight = importance = gradient magnitude  
\end{abstract}

\vspace{-0.2cm}
\section{Introduction}
\label{introduction}
\vspace{-0.2cm}

In this work, we target at robust deep learning, which is indispensable when it comes to large-scale industrial applications. 
%
%Due to the availability of vast training datasets and powerful computation resources, deep learning has achieved great success in diverse tasks, e.g., computer vision \cite{krizhevsky2012imagenet} and speech recognition \cite{hinton2012deep}. 
%and reinforcement learning \cite{silver2016mastering}. 
%However, 
It is non-affordable to guarantee the quality of training data as its scale grows dramatically. Consequently, abnormal examples\footnote{A training example is denoted as an observation-label pair, where the observation can be an image or video while the label defines its semantic information. We regard a training example as abnormal unrestrictedly whenever its observation and label are semantically unmatched, e.g., out-of-distribution examples (the observations contain only background or objects that do not belong to any training class), or examples with wrongly annotated labels.} generally exist in large-scale real-world scenarios \cite{berrada2018smooth}, which is caused by many factors, such as incomplete annotation, wrong labelling, subjectiveness, bias and so forth. 
%We show some abnormal examples in Figure~\ref{fig:abnormal_examples}.
%Therefore, learning robust feature embeddings is fundamental and significant in deep learning tasks.  
%Therefore, it is significantly challenging to learn robust deep neural networks (DNNs) against overfitting to examples with noisy labels. 
Unfortunately, DNNs trained with categorical cross entropy (CCE) can fit random patterns \cite{zhang2017understanding}.

Great advances have been made towards training DNNs robustly when abnormal training examples exist \cite{arpit2017closer,chang2017active,ren2018learning,jiang2018mentornet}. The robust loss function is one of them. In this paper, we study a so-claimed robust loss function, mean absolute error (MAE) following \cite{ghosh2017robust, zhang2018generalized}. 
%It is complementary to prior work and can inspire future research towards robust DNNs against arbitrary abnormal training examples.
%
%Concretely, 
According to the theoretical analysis of CCE and MAE in \cite{ghosh2017robust}, CCE is sensitive to label noise while MAE is noise-tolerant. Thereafter, generalised cross entropy (GCE) \cite{zhang2018generalized} concludes MAE treats training samples equally, thus being noise-robust. 

\begin{figure*}[!t]
	\centering
	\begin{minipage}[t]{0.51\textwidth}
		\vspace{0.3cm}
		\centering
		\captionof{table}{
			Classification accuracy (\%) of CCE, MAE, and IMAE on CIFAR-10 \cite{krizhevsky2009learning}. 
			40\% of training examples, i.e., the noisy subset, have wrong labels. 
			We test each model's performance on test set, noisy subset and clean subset of training data. 
			The backbone is ResNet56 owning enough capacity \cite{he2016deep}. 
		}%
		\setlength{\tabcolsep}{2.5pt} % Default value: 6pt
		\fontsize{7.7pt}{7.7pt}\selectfont
		\begin{tabular}{lccc}
			\toprule
			Loss &  \makecell{Test set\\(Generalisation)} & \makecell{Noisy subset\\(Noise-tolerance)} & \makecell{Clean subset\\(Learning ability)}\\
			\midrule
			%		\multirow{3}{*}{ResNet20}
			%		& CCE & 67.0 & 34.3 & 93.3\\
			%		& MAE & 75.9 & 6.8 & 84.6\\
			%		& IMAE & 84.0 & 5.5 & 94.0\\
			%\multirow{3}{*}{ResNet56}
			CCE & 63.3 & 75.0 & 96.2\\
			MAE & 66.9 & 8.1 & \textbf{74.3} (worst)\\
			\midrule
			\makecell{IMAE} & \textbf{81.5} (best) & \textbf{6.5} (best) & 93.1\\
			\bottomrule
		\end{tabular}
		\label{table:MAE_observations}
	\end{minipage}
	\hfill
	\begin{minipage}[t]{0.48\textwidth}
		\centering
		\vspace{-0.5cm}
		\includegraphics[width=0.66\textwidth,valign=t]{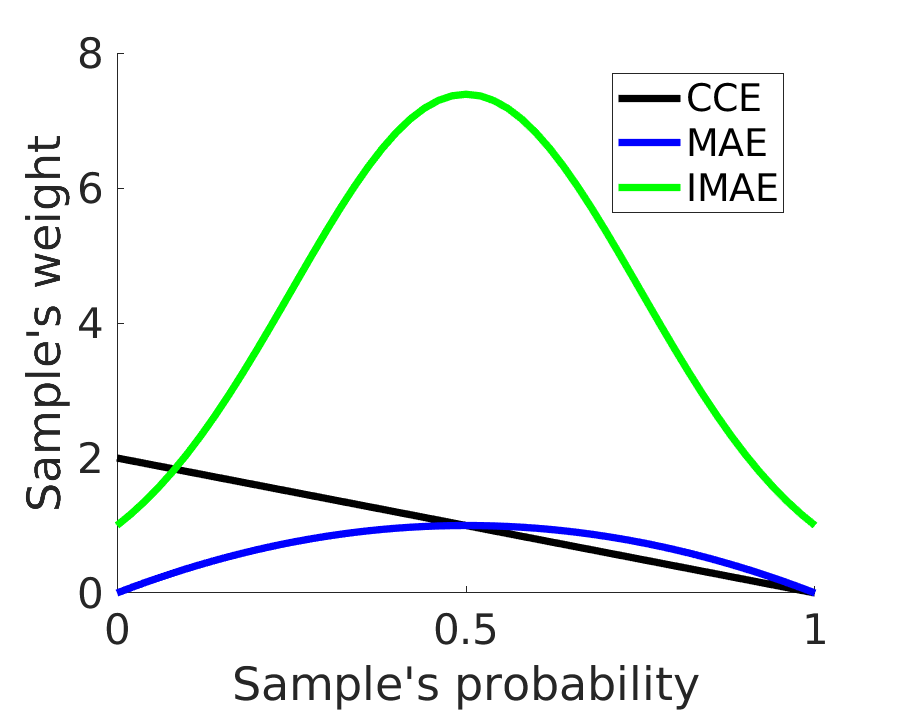}
		\vspace{-0.2cm}
		\captionof{figure}{
			Sample's weight along with sample's probability being classified to its labelled class in CCE, MAE, IMAE with $T=8$. 
			%CCE prioritises low-probability examples to which abnormal ones generally belong.
			%MAE and IMAE own better robustness by emphasising on medium-probability ones.
			%
			If probabilities are uniformly distributed, the variances of CCE's, MAE's and IMAE's weighting curves are 0.33, 0.09 and 4.55, respectively. 
		}
		\label{fig:absolute_weight_CCE_MAE_IMAE_V02}
	\end{minipage}%
	\vspace{-0.4cm}
\end{figure*}

However, our empirical observation and technical analysis lead us to a contradictory and more reasonable conclusion. 
\\
\textit{Observation: In Table~\ref{table:MAE_observations}, when 40\% noise exists, compared with CCE, MAE underfits to clean training data points, thus fitting much fewer abnormal examples. }
\\
\textit{Conclusion: In Figure~\ref{fig:absolute_weight_CCE_MAE_IMAE_V02}, MAE emphasises more on uncertain examples, whose probabilities of being classified to its labelled class are around 0.5, thus being noise-robust. }  

Specifically, according to Table~\ref{table:MAE_observations}, MAE is much more noise-tolerant than CCE. However, its ability of learning meaningful patterns is much weaker, fitting only 74.3\% of the clean subset. We provide an intuitive interpretation for this according to Figure~\ref{fig:absolute_weight_CCE_MAE_IMAE_V02}:
\textit{The variance of MAE's weight curve along with probability is only 0.09. As a result, the impact ratio between two examples is too small}.\footnote{The terms, examples' weight or impact, and examples' gradient magnitude w.r.t. logits, are used interchangeably because we define the weight by gradient's magnitude. The impact ratio between two examples is changed only when gradients' magnitude is scaled non-linearly.}
% as shown in Figure~\ref{fig:absolute_weight_CCE_MAE_IMAE_V02}. 
The impact ratio reflects the relative impact of one example versus another for updating parameters.
%, which can be measured by the weight variance of all examples. 
Due to MAE's small weight variance, informative samples cannot contribute enough against non-informative ones. 
Therefore, MAE cannot learn meaningful patterns well and is not widely used.

%Secondly, 
%Based on our interpretation, 
%Our second contribution is
To adjust MAE's weight variance, we design an effective and simple solution, IMAE, which non-linearly transforms MAE's weighting scheme by an exponential function. 
%IMAE makes the differentiation degree over training samples controllable by choosing a proper exponentiation base. 
On the one hand, by preserving MAE's overall weighting scheme, IMAE is noise-robust. On the other hand, by making the gradient magnitude's variance over training examples controllable, it learns meaningful patterns much better.
%
%Thirdly, 
%We present extensive empirical studies of CCE, MAE, and our IMAE on two tasks: image classification and video-based person re-identification. 
%Regarding image classification on CIFAR-10 \cite{krizhevsky2009learning}, we demonstrate
%the effectiveness of our IMAE against CCA and MAE
%consistently when the noise rate ranges from 0\% to 80\%. As noise rate increases, the superiority of IMAE becomes more noticeable. 
%Moreover, it consolidates our analysis about the data fitting issues of CCE (overfitting), MAE (underfitting) and IMAE (proper fitting) in Sec.~\ref{sec:improved_MAE}.
%In video-based person re-identification \cite{zheng2016mars} where noise exists but noise rate is unknown, and test classes are unseen during training, our IMAE also outperforms CCE, MAE significantly (Sec.~\ref{sec:experiments}). 

We demonstrate the effectiveness of IMAE under different scenarios.
%Especially when noise rate increases, the superiority of IMAE becomes more noticeable. 
%Regarding video retrieval, different noise types exist and the noise rate is unknown. Moreover, testing classes are unseen during training. IMAE also outperforms CCE, MAE and GCE significantly. 
%
Most importantly, these empirical evidences justify that our interpretation of MAE's underfitting problem is reasonable and our proposed solution is superior. %It is non-trivial to prove theoretically which weighting scheme is better under real-world scenarios as DNNs are capable of fitting abnormal training examples perfectly \cite{zhang2017understanding}.
Our key findings are summarised as follows: 
\begin{itemize}
\vspace{-0.2cm}
\item CCE overfits to noise easily because it emphasises on low-probability examples to which abnormal ones generally belong. Although CCE's weight variance is not large (0.33), its fitting ability benefits from emphasising on low-probability examples. 

\vspace{-0.2cm}
\item MAE is noise-robust by focusing on uncertain (medium-probability) examples instead of treating all equally.  However, MAE generally underfits due to its small weights variance (0.09), leading to small impact ratio between even far different examples.

\vspace{-0.2cm}
\item Our proposed IMAE achieves new state-of-the-art on robust training against synthetic label noise and realistic unknown noise simply by adjusting MAE's weight variance, which is inspiring.
\vspace{-0.2cm}
\end{itemize}

%%%%%%%%%%%%%%%%%%%%%%%%%%%%%%%%%%%%%%%%%%%%%%%%%%%%%
%%%%%%%%%%%%%%%%%%%%%%%%%%%%%%%%%%%%%%%%%%%%%%%%%%%%

%%%%%%%%%%%%%%%%%%%%%%%%%%%%%%%%%%%%%%%%%%%%%%%%%%%%%
%%%%%%%%%%%%%%%%%%%%%%%%%%%%%%%%%%%%%%%%%%%%%%%%%%%%

\vspace{-0.2cm}
%start-------------------------------------------------------------------------
\section{Preliminaries}
\vspace{-0.2cm}

We denote a training mini-batch as $\mathbf{X}=\{(\mathbf{x}_i, y_i)\}_{i=1}^{N}$, where there are $N$ samples. $(\mathbf{x}_i, y_i)$ represents $i$-th training sample $\mathbf{x}_i \in \mathbb{R}^D$ and its annotated class label $y_i \in \{1,2,...,C\}$. $D$ is the dimensionality of input samples and $C$ is the number of all training classes.
Let $f_{\theta}$ be a deep neural network, which transforms $\mathbf{x}_i$ to a representation $\mathbf{f}_i = f_{\theta}(\mathbf{x}_i) \in \mathbb{R}^E$, $E$ is the dimensionality of target space and $\theta$ indicates the parameters to be learned. %To learn robust deep embeddings, 
%In this work, $f_{\theta}$ is .
%which can be designed as different architectures (a.k.a. backbones) in different applications. 

To optimise $f_{\theta}$ during training, a linear classifier is generally trained jointly \cite{liu2016large}. In general, the linear classifier follows the output embeddings and is composed of one $C$-neuron fully connected (FC) layer, one softmax normalisation layer and one loss layer. The FC layer can be represented as $\mathbf{z}_i = \mathbf{W}^\top \mathbf{f}_i \in \mathbb{R}^C$, where $\mathbf{W} = [\mathbf{w}_1, \mathbf{w}_2, ..., \mathbf{w}_C] \in \mathbb{R}^{E\times C}$ consists of $C$ weight vectors (the bias term is omitted for brevity). $\mathbf{z}_{ij} = \mathbf{w}_j^\top \mathbf{f}_i$ is a logit which indicates the compatibility between sample $\mathbf{x}_i$ and class $j$. To produce the probabilities of sample $\mathbf{x}_i$ belonging to different classes, we normalise its logit vector $\mathbf{z}_i$ using a softmax function: 
%\begin{equation}
%\label{eq:probability}
$p(j|\mathbf{x}_i) = \frac{\exp(\mathbf{z}_{ij})}{\sum_{m=1}^{C} \exp(\mathbf{z}_{im})} ,$
%\end{equation} 
where $p(j|\mathbf{x}_i)$ is the probability of sample $\mathbf{x}_i$ being predicted to class $j$. 

Let $q(j|\mathbf{x}_i)$ be the ground-truth probability of $\mathbf{x}_i$ belonging to class $j$, i.e., $q(j|\mathbf{x}_i)=1$ if $j=y_i$, $q(j|\mathbf{x}_i)=0$ otherwise. In the loss layer, if we use CCE, the minimisation objective per iteration is: 
\vspace{-0.2cm}
\begin{equation}
\fontsize{9pt}{9pt}\selectfont
\label{eq:loss_CCE}
\begin{aligned}
	L_{\mathrm{CCE}} (\mathbf{X};f_\theta,\mathbf{W}) 
	%&= 	 \frac{1}{N} \sum_{i=1}^N L_{\mathrm{CCE}} (\mathbf{x}_i;f_\theta,\mathbf{W})\\
	%&= 	- \frac{1}{N} \sum_{i=1}^N \sum_{j=1}^{C} \log(p(j|\mathbf{x}_i))q(j|\mathbf{x}_i) \\
	= 	- \frac{1}{N} \sum_{i=1}^N \sum_{j=1}^{C} q(j|\mathbf{x}_i) \log p(j|\mathbf{x}_i) 
	= 	
	%- \frac{1}{N} \sum_{i=1}^N \log(p(y_i|\mathbf{x}_i))
	- \frac{1}{N} \sum_{i=1}^N \log p(y_i|\mathbf{x}_i)
	. 
\end{aligned}
\end{equation}
If MAE is applied, the minimisation objective becomes: 
\vspace{-0.2cm}
\begin{equation}
\fontsize{9pt}{9pt}\selectfont
\label{eq:loss_MAE}
\begin{aligned}
	L_{\mathrm{MAE}} (\mathbf{X};f_\theta,\mathbf{W}) 
	%&= 	\frac{1}{N} \sum_{i=1}^N L_{\mathrm{MAE}} (\mathbf{x}_i;f_\theta,\mathbf{W}) \\
	= 	 \frac{1}{N} \sum_{i=1}^N \sum_{j=1}^{C}
	|p(j|\mathbf{x}_i)- q(j|\mathbf{x}_i) |
	= 	 \frac{2}{N} \sum_{i=1}^N 
	(1-p(y_i|\mathbf{x}_i))
	,
\end{aligned}
%\vspace{-0.2cm}
\end{equation}
where $|\cdot|$ is the absolute function. 
%
%As demonstrated above, when learning deep representations with the supervision of CCE or MAE, our objective is to search the optimal parameters $\theta$ and $\mathbf{W}$ for the embedding network and linear classifier, respectively.

\begin{figure*}[t]
	\centering
	\begin{minipage}[t]{0.495\textwidth}
		\centering
		\includegraphics[width=0.999\linewidth,valign=t]{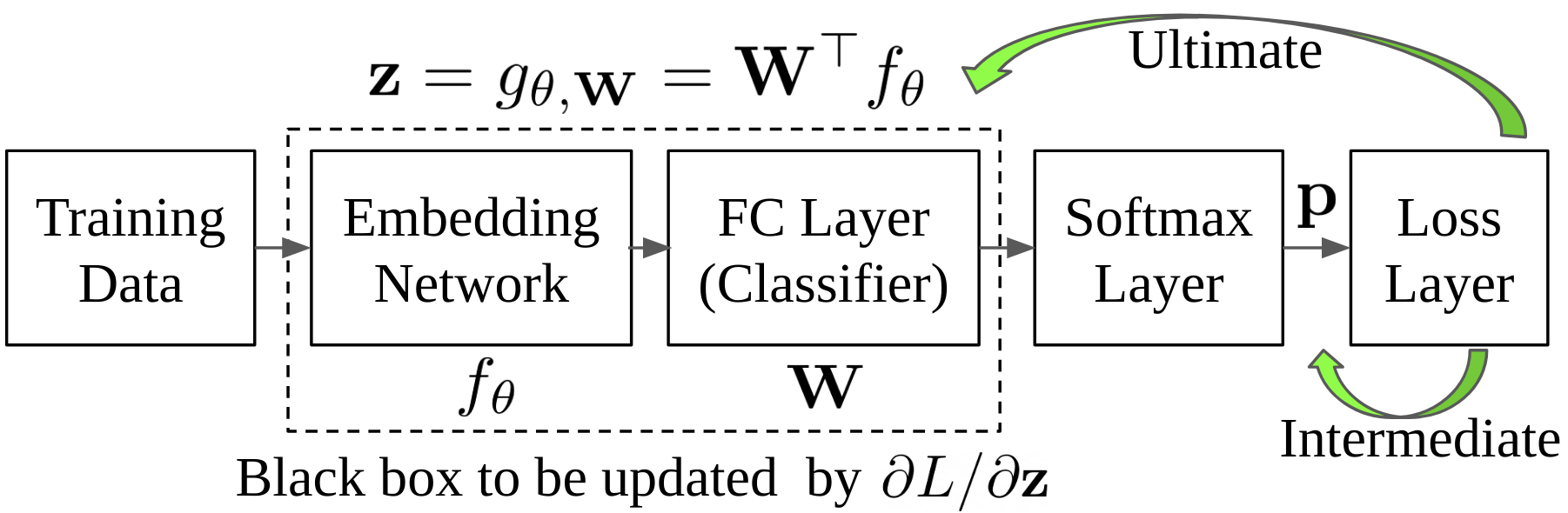}
		\captionof{figure}{
			Pipeline of a softmax deep network. There are two reasons for analysing loss functions based on $\frac{\partial L}{\partial \mathbf{z}}$: (1) 
			%In the deep learning toolboxes, e.g., Caffe \cite{jia2014caffe}, 
			In gradient back-propagation, the gradients of examples in a mini-batch are fused when computing $\frac{\partial L}{\partial \mathbf{z}}$.  
			(2) 
			Intermediate differences of $\frac{\partial L}{\partial \mathbf{p}}$ lead to ultimate differences of $\frac{\partial L}{\partial \mathbf{z}}$.
			%\cite{zhang2018generalized} analysed the differences of MAE and CCE according to $\frac{\partial L}{\partial \mathbf{p}}$
			Therefore, our analysis of $\frac{\partial L}{\partial \mathbf{z}}$ is more direct versus that of $\frac{\partial L}{\partial \mathbf{p}}$ in  \cite{zhang2018generalized}.
			%The gradient computations $\frac{\partial \mathbf{z}}{\partial \mathbf{W}}$ and $\frac{\partial \mathbf{z}}{\partial{\theta}}$ are the same for any example; 
		}
		\label{fig:embedding_pipeline}
	\end{minipage}
	\hfill
	\begin{minipage}[t]{0.46\textwidth}
		\centering
		\vspace{-0.5cm}
		\includegraphics[width=0.91\textwidth,valign=t]{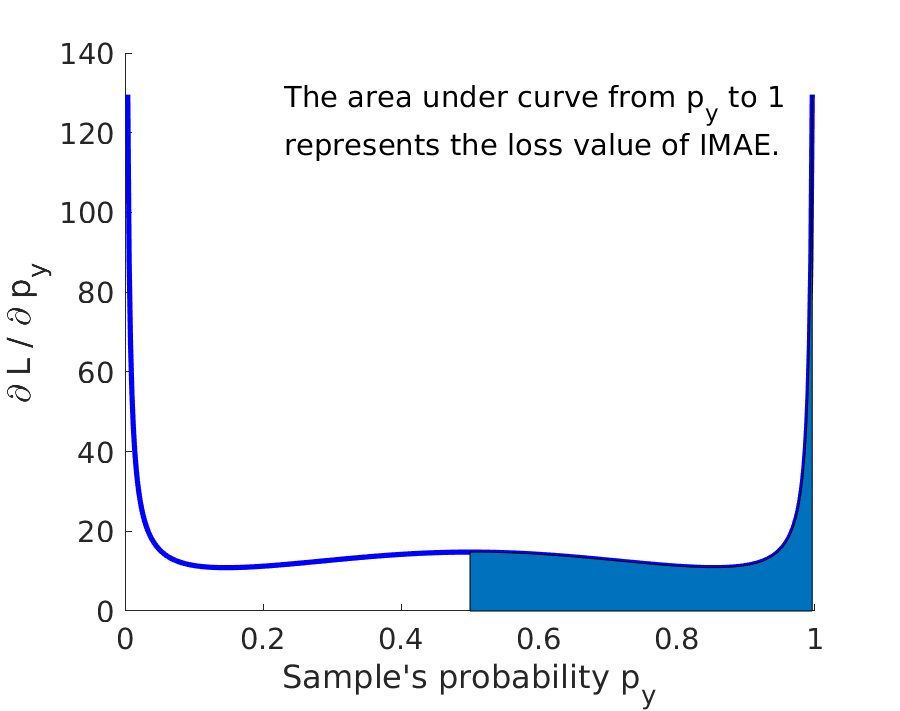}
		\vspace{-0.2cm}
		\captionof{figure}{
			Although the loss expression of IMAE is not an elementary function, we visualise it by integral, i.e., the area under curve from $p_y$ to 1.    
		}
		\label{fig:loss_visualisation_IMAE}
	\end{minipage}%
	\vspace{-0.1cm}
\end{figure*}

In summary, we learn a softmax deep network $g_{\theta, \mathbf{W}}$, which outputs logits: $\mathbf{z}_i = g_{\theta, \mathbf{W}}(\mathbf{x}_i)=\mathbf{W}^\top f_{\theta}(\mathbf{x}_i) \in \mathbb{R}^C.$
%\begin{equation}
%	\mathbf{z}_i = g_{\theta, \mathbf{W}}(\mathbf{x}_i)=\mathbf{W}^\top f_{\theta}(\mathbf{x}_i) \in \mathbb{R}^C
%	.
%\end{equation}
In classification tasks, 
%\cite{krizhevsky2009learning,krizhevsky2012imagenet,he2016deep}
we use $\mathbf{z}=g_{\theta, \mathbf{W}}(\mathbf{x})$ to produce logits for a test image $\mathbf{x}$. 
%We can also normalise the logits using a softmax function to obtain the probabilities of a test image being predicted to different classes. 
While in verification or retrieval tasks \cite{wang2019ranked,wang2019deep,wang2019id}, we only use 
%the penultimate layer's output of $g_{\theta, \mathbf{W}}$, i.e., 
$\mathbf{f}=f_{\theta}(\mathbf{x})$ as an embedding function. 
%Generally, features are first $L_2$ normalised and their similarities are then measured by their cosine distances. 
%
The overall pipeline is described in Figure~\ref{fig:embedding_pipeline}.
The output of the softmax layer is $\mathbf{p}$.
%%%%%%%%%%%%%%%%%%%%%%%%%%%%%%%%%%%%%%%%%%%%%%%%%%%%%

%%%%%%%%%%%%%%%%%%%%%%%%%%%%%%%%%%%%%%%%%%%%%%%%%%%%
%end-------------------------------------------------------------------------

\textbf{Definition 1} (Uncertain Examples). \textit{We define uncertain examples to be those data points whose $p(y_i|\mathbf{x}_i)$ are around 0.5. Given an example $\mathbf{x}_i$, if its $p(y_i|\mathbf{x}_i)$ is closer to 0.5, its uncertainty is higher.}
\\
\textbf{Remark 1}. 
This definition of uncertain examples is intuitive. \textit{If $p(y_i|\mathbf{x}_i)$ is closer to 1, the confidence of $\mathbf{x}_i$ being class $y_i$ is higher. If $p(y_i|\mathbf{x}_i)$ is closer to 0, the confidence of $\mathbf{x}_i$ belonging to one of other classes is higher. However, if $p(y_i|\mathbf{x}_i)$ is around 0.5, we are more uncertain about whether $\mathbf{x}_i$ being class $y_i$.}
Therefore, we can understand uncertainty from the perspective of binary classification (Logistic Regression), i.e., whether $\mathbf{x}_i$ being class $y_i$ or not. 
\\
\textbf{Remark 2}. 
We have \textit{the premise that abnormal (noisy) examples have smaller probabilities in general}. This premise is widely used and demonstrated by our empirical observations. For example, in Figure~\ref{fig:ResNet2056_Noise40_CCE_MAE_SMAE_dynamics} and Tables~\ref{table:MAE_observations},~\ref{table:CIFAR_noise_three}, the accuracy of noisy subset is less than that of clean subset consistently.    
\\
\textbf{Remark 3}. The uncertainty of an example is determined by its probability of being classified to its annotated label. This example can belong to one of the training classes (uncertain in-distribution example), or a class which does not exist in the training set (uncertain out-of-distribution example).

\vspace{-0.2cm}
%start-------------------------------------------------------------------------
\section{Gradient Magnitude Serving as Weight}
\label{sec:gradient_assignment_analysis}
\vspace{-0.2cm}

%We would like to analyse samples' contribution towards searching optimal network parameters from the perspective of gradient assignment. 
As shown in Figure~\ref{fig:embedding_pipeline}, $g_{\theta, \mathbf{W}}$ can be viewed as a black box and the update of $\theta$ and $\mathbf{W}$ is based on the back-propagation of logits' gradient.
%$\mathbf{z}$.
%\triangledown  $\partial L/ \partial \mathbf{z}$
Therefore, an example's contribution can be measured by the magnitude of its partial derivative w.r.t. $\mathbf{z}$. 
%This can also be regarded the weighting of samples that is embedded implicitly in a loss function. 
%Understanding the implicit weighting of samples in a loss function is not well studied yet. 
It can be regarded as example weighting that is naturally built-in in loss functions. 
%and not well studied yet. 
%\\
%
%We focus on two well-known losses, CCE and MAE, motivated by \cite{ghosh2017robust,zhang2018generalized,rooyen2015learning}. 
%\\
%By analysing their sample weighting schemes, we present a new understanding of their behaviours in practice: CCE is prone to overfit the training data while MAE generally underfits it, which is also observed in Section~\ref{sec:experiments}. 
%
%In this work, we focus on the implicit weighting in two well-known losses, CCE and MAE. Based on our implicit sample weighting analysis of CCE and MAE from the perspective of gradient assignment, we obtain differnt conclusions from \cite{ghosh2017robust}, which interests us a lot. 
%
For brevity and clarity, we summarise the results here and put the detailed derivation in our supplementary material. 

\vspace{-0.2cm}
\subsection{Derivation of Softmax, CCE and MAE Layers}
\vspace{-0.2cm}
%As the softmax layer is shared by CCE and MAE, we present the derivation of softmax layer first. 

\noindent
According to $p(j|\mathbf{x}_i)$, Eq.~(\ref{eq:loss_CCE}), 
Eq.~(\ref{eq:loss_MAE}), 
we have\\
\begin{minipage}{\textwidth}
\begin{minipage}{.43\textwidth}
	\fontsize{7.8pt}{7.8pt}\selectfont
	\begin{equation}
		\label{eq:both_prob_derivation_final_MainPaper}
		\begin{aligned}
			\frac{\partial p(y_i|\mathbf{x}_i)}{\partial \mathbf{z}_{ij}} 
			&=
			\begin{cases} 
				p(y_i|\mathbf{x}_i) 
				(1-p(y_i|\mathbf{x}_i))
				\text{, } &j = y_i  \\
				-p(y_i|\mathbf{x}_i) 
				p(j|\mathbf{x}_i)  
				\text{, } &j \neq y_i
			\end{cases};
		\end{aligned}
	\end{equation}
\end{minipage}%
\hfill
\begin{minipage}{.34\textwidth}
	\fontsize{7.5pt}{7.5pt}\selectfont
	\begin{equation}
		\label{eq:derivation_CCE_x_i_MainPaper}
		\begin{aligned}
			\frac{\partial L_{\mathrm{CCE}}(\mathbf{x}_i)}{\partial p(j|\mathbf{x}_i)} 
			&=
			\begin{cases} 
				%\frac{-1}{p(y_i|\mathbf{x}_i)} 
				-p(y_i|\mathbf{x}_i)^{-1} 
				\text{, } &j = y_i  \\
				0       
				\text{, } &j \neq y_i
			\end{cases};
		\end{aligned}
	\end{equation}
\end{minipage}
\hfill
\begin{minipage}{0.21\textwidth}
	\fontsize{7.5pt}{7.5pt}\selectfont
	\begin{equation}
		\label{eq:derivation_MAE_x_i_MainPaper}
		\begin{aligned}
			\frac{\partial L_{\mathrm{MAE}}(\mathbf{x}_i)}{\partial p(j|\mathbf{x}_i)} 
			&=
			\begin{cases} 
				-2  \text{, } &j = y_i  \\
				0        \text{, } &j \neq y_i
			\end{cases}
			.
		\end{aligned}
	\end{equation}
\end{minipage}
\end{minipage}

\vspace{-0.2cm}
\subsection{Perspective of Derivatives w.r.t. Logits Other Than Probabilities}
\vspace{-0.2cm}

\textit{Prior conclusion according to $\frac{\partial L_{\mathrm{CCE}}(\mathbf{x}_i)}{\partial p(j|\mathbf{x}_i)}$, $\frac{\partial L_{\mathrm{MAE}}(\mathbf{x}_i)}{\partial p(j|\mathbf{x}_i)}$}:
\cite{zhang2018generalized} concludes that CCE is sensitive to abnormal examples while MAE is robust by treating all data points equally according to Eq.~(\ref{eq:derivation_CCE_x_i_MainPaper}) and Eq.~(\ref{eq:derivation_MAE_x_i_MainPaper}), respectively.
%
%
%Instead, we propose to understand more reasonably and better from the angle of derivatives w.r.t. logits. 
%Our reason is that logit vector $\mathbf{z}_{i}$ is the exact output of all parametric layers, and the update of parameters is based on its gradient back-propagation. 

In this work, we propose to further analyse $\frac{\partial L_{\mathrm{CCE}}(\mathbf{x}_i)}{\partial \mathbf{z}_{ij}}$, $\frac{\partial L_{\mathrm{MAE}}(\mathbf{x}_i)}{\partial \mathbf{z}_{ij}}$ as discussed in Figure~\ref{fig:embedding_pipeline}. 
\noindent
According to Eq.~(\ref{eq:both_prob_derivation_final_MainPaper}), ~(\ref{eq:derivation_CCE_x_i_MainPaper})  and~(\ref{eq:derivation_MAE_x_i_MainPaper}), we calculate: 
% $\partial L_{\mathrm{CCE}}(\mathbf{x}_i)/ \partial \mathbf{z}_{i}$:
\begin{minipage}{\textwidth}
	\vspace{-0.2cm}
	\begin{minipage}{0.435\textwidth}
		\fontsize{8pt}{8pt}\selectfont
		\begin{equation}
			\label{eq:summary_CCE_z_MainPaper}
			\begin{aligned}
				\frac{\partial L_{\mathrm{CCE}}(\mathbf{x}_i)}{\partial \mathbf{z}_{ij}} 
				&=
				\begin{cases} 
					p(y_i|\mathbf{x}_i)-1  \text{, } &j = y_i  \\
					p(j|\mathbf{x}_i)        \text{, } &j \neq y_i
				\end{cases}
				.
			\end{aligned}
		\end{equation}
	\end{minipage}%
	\hfill
	\begin{minipage}{0.54\textwidth}
		\fontsize{8.5pt}{8.5pt}\selectfont
		\begin{equation}
			\label{eq:summary_MAE_z_MainPaper}
			\begin{aligned}
				\frac{\partial L_{\mathrm{MAE}}(\mathbf{x}_i)}{\partial \mathbf{z}_{ij}} 
				&=
				\begin{cases} 
					2 p(y_i|\mathbf{x}_i)
					(p(y_i|\mathbf{x}_i)-1) \text{,} &j = y_i  \\
					2 p(y_i|\mathbf{x}_i)
					p(j|\mathbf{x}_i) \text{,} &j \neq y_i
				\end{cases}
				.
			\end{aligned}
		\end{equation}
	\end{minipage}
\end{minipage}

%\noindent
%Analogously, according to Eq.~(\ref{eq:derivation_MAE_x_i_MainPaper}) and~(\ref{eq:both_prob_derivation_final_MainPaper}), we have:

\noindent
\textbf{{Gradient magnitude treated as weight.}}
In CCE and MAE, training samples are weighted because different ones own different gradient magnitude w.r.t. logit vector $\mathbf{z}$.
We choose to measure one gradient's magnitude by its $L_1$ norm because of its simpler statistics than other norms. If one sample's gradient is larger, its impact is larger during gradient back-propagation. 
%In other words, sample's weight value equals to its gradient magnitude w.r.t. $\mathbf{z}_{i}$. 

For CCE, based on Eq.~(\ref{eq:summary_CCE_z_MainPaper}), the weight of sample $\mathbf{x}_{i}$ is:
\begin{equation}
	\fontsize{9pt}{9pt}\selectfont
\label{eq:implicit_CCE}
w_\mathrm{CCE}(\mathbf{x}_{i}) = ||\frac{\partial L_{\mathrm{CCE}}(\mathbf{x}_i)}{\partial \mathbf{z}_{i}}||_1 = 2
%\times
(1-p(y_i|\mathbf{x}_i)),
\end{equation}
where $||\cdot||_1$ denotes $L_1$ norm. 
For MAE, based on Eq.~(\ref{eq:summary_MAE_z_MainPaper}), the weight of sample $\mathbf{x}_{i}$ is:
\begin{equation}
	\fontsize{9pt}{9pt}\selectfont
\label{eq:implicit_MAE}
w_\mathrm{MAE} (\mathbf{x}_{i}) = ||\frac{\partial L_{\mathrm{MAE}}(\mathbf{x}_i)}{\partial \mathbf{z}_{i}}||_1 = 4
%\times
p(y_i|\mathbf{x}_i) 
%\times
(1-p(y_i|\mathbf{x}_i)).
\end{equation}
According to Eq.~(\ref{eq:implicit_CCE}) and Eq.~(\ref{eq:implicit_MAE}), in both CCE and MAE, examples' impact is determined by their probabilities being predicted to annotated labels. 
%Interestingly, this is different from the conclusion that MAE treats samples equally \cite{ghosh2017robust,zhang2018generalized}. We find that it is more reasonable to understand the implicit weighting from the derivatives w.r.t. $\mathbf{z}_{i}$ instead of $p(j|\mathbf{x}_i)$ in \cite{ghosh2017robust,zhang2018generalized} because  $\mathbf{z}_{i}$ is the exact output of all parametric layers as shown in Figure~\ref{fig:embedding_pipeline}.
%\\
%The comparison between CCE and MAE is given in Sec.~\ref{sec:improved_MAE} together with our proposal, improved MAE (IMAE).
%end-------------------------------------------------------------------------

\vspace{-0.2cm}
%start------------------------------------------------------------------------
\section{Improved MAE}
\label{sec:improved_MAE}
\vspace{-0.2cm}

IMAE transforms MAE's weighting scheme non-linearly:
\begin{equation}
	\fontsize{9pt}{9pt}\selectfont
w_\mathrm{IMAE} (\mathbf{x}_{i}) =  \exp(T   
%\times
p(y_i|\mathbf{x}_i) 
%\times
(1-p(y_i|\mathbf{x}_i))),
\label{eq:weight_IMAE}
\end{equation}
where $T$ controls the exponential base. 
%Note that the improved MAE is derived from the perspective of $L_1$ norm w.r.t. $\mathbf{z}_i$ instead of loss value. 
%We remark that the loss computation of MAE is unchanged. %Gradient matters. 
In back-propagation, we simply scale the gradient w.r.t. logits as follows: 
\begin{equation}
	\fontsize{9pt}{9pt}\selectfont
\label{eq:implicit_IMAE}
\begin{aligned}
	&\frac{\partial L_{\mathrm{IMAE}}(\mathbf{x}_i)}{\partial \mathbf{z}_{i}} = \frac{\partial L_{\mathrm{MAE}}(\mathbf{x}_i)}{\partial \mathbf{z}_{i}}  \frac{w_\mathrm{IMAE} (\mathbf{x}_{i})}{w_\mathrm{MAE} (\mathbf{x}_{i})}
	%\\
	%&
	\Rightarrow
	||\frac{\partial L_{\mathrm{IMAE}}(\mathbf{x}_i)}{\partial \mathbf{z}_{i}}||_1 = w_\mathrm{IMAE} (\mathbf{x}_{i}).
\end{aligned}
\end{equation}

IMAE is a family of robust losses when $T$ changes, as summarised in Table~\ref{table:GR_differentiation} and Figure~\ref{fig:loss_visualisation_IMAE}.

\noindent
\subsection{Design Motivation: To Adjust Gradient Magnitude's Variance and Impact Ratio}
%
%IMAE transforms MAE's weighting scheme using an exponential function, where a scaling parameter $T$ adjusts the weighting variance, i.e., differentiation degree of training samples. 
%
Linear scaling also changes magnitude variance. However, it cannot adjust impact ratio, i.e., the ratio between two gradients' magnitude. That is why we have tried linear scaling and find it does not work. 

Instead, the exponential function is non-linear so that the impact ratio of one sample versus another is re-adjusted compared with original MAE. 
%
%Moreover, 
The hyper-parameter $T$ controls how significant gradient magnitude's variance and impact ratio are changed. 

Furthermore, assuming that samples' probabilities are uniformly distributed, we compute the gradients' variance of MAE and IMAE over training data points: 
\begin{equation}
	\begin{aligned}
		\sigma_{\mathrm{MAE}}&=
		\displaystyle\int^{1}_{0} 
		w_{\mathrm{MAE}}^{2} (p)
		\, \mathrm d p
		- (\displaystyle\int^{1}_{0} 
		w_{\mathrm{MAE}} (p)
		\, \mathrm d p)^2
	\end{aligned}
\end{equation}
%\vspace{-0.6cm}
\begin{equation}
	\label{eq:variance_IMAE}
	\begin{aligned}
		\sigma_{\mathrm{IMAE}}&=
		\displaystyle\int^{1}_{0} 
		w_{\mathrm{\mathrm{IMAE}}}^{2} (p)
		\, \mathrm d p
		- (\displaystyle\int^{1}_{0} 
		w_{\mathrm{IMAE}} (p)
		\, \mathrm d p)^2.
	\end{aligned}
\end{equation}
We have $\sigma_{\mathrm{MAE}}=0.09$. When $T=8$, $\sigma_{\mathrm{IMAE}} = 4.55$.

%During training, absolute weight value is not informative enough and is usually scaled by learning rates, or regularisation techniques like weight decay. 
%We find that relative weight between samples is helpful for understanding how different samples are differentiated.
%Therefore, instead of understanding merely from absolute weight $w_\mathrm{CCE}(\mathbf{x}_{i}), w_\mathrm{MAE}(\mathbf{x}_{i}), w_\mathrm{IMAE}(\mathbf{x}_{i})$, we calculate the relative weight between two samples for each loss:
%\begin{equation}
%	\label{eq:relative_CCE}
%	\frac{w_\mathrm{CCE}(\mathbf{x}_{i})}{w_\mathrm{CCE}(\mathbf{x}_{j})}  = 
%	\frac{ 1-p(y_i|\mathbf{x}_i)}
%	{ 1-p(y_j|\mathbf{x}_j)}.
%\end{equation}
%\begin{equation}
%	\label{eq:relative_MAE}
%	\frac{w_\mathrm{MAE}(\mathbf{x}_{i})}{w_\mathrm{MAE}(\mathbf{x}_{j})}  = 
%	\frac{ 
	%		p(y_i|\mathbf{x}_i) \times (1-p(y_i|\mathbf{x}_i)) 
	%	}
%	{ 
	%		p(y_j|\mathbf{x}_j) \times (1-p(y_j|\mathbf{x}_j))
	%	}.
%\end{equation}
%\begin{equation}
%	\label{eq:relative_IMAE}
%	\frac{w_\mathrm{IMAE}(\mathbf{x}_{i})}{w_\mathrm{IMAE}(\mathbf{x}_{j})}  = 
%	\frac
%	{\exp(T\times p(y_i|\mathbf{x}_i) \times (1-p(y_i|\mathbf{x}_i)))}
%	{\exp(T\times p(y_j|\mathbf{x}_j) \times (1-p(y_j|\mathbf{x}_j)))}
%	.
%\end{equation}

\noindent
\subsection{Discussion of MAE and CCE}
The weighting curves of CCE, MAE and IMAE are compared in Figure \ref{fig:absolute_weight_CCE_MAE_IMAE_V02}.  
Our key findings are summarised in the end of introduction. 
We further discuss them as follows:

\begin{itemize}
	%\vspace{-0.2cm}
	%\item CCE emphasises on samples with lower probabilities, to which abnormal examples generally belong. Therefore, \textit{CCE is prone to overfitting} in practice. 
	
	\vspace{-0.3cm}
	\item %MAE assigns low weights to examples with low probabilities, thus being noise-tolerant.
	{MAE's weighting scheme is appealing and practical} 
	in that samples with medium probabilities are emphasized.
	Generally, high-probability samples are clean and already trained well. 
	While low-probability ones are highly likely to be noisy as a model improves during training. 
	Although all samples are not trained well and probabilities are not meaningful at the beginning, it also does not hurt to focus on medium-probability ones.    
	\item {MAE's gradient magnitude's variance over data points is only 0.09}. As a consequence, the impact ratio of one example versus another is too small. Therefore, the majority contribute almost equally. {Therefore, MAE generally underfits to  training data.} 
	
	\vspace{-0.2cm}
	\item Does high loss value usually back-propagate high gradients to update parameters? The answer is NO. Therefore, those theorems based on loss values, e.g., symmetric or bounded conditions are insufficient for analysing robustness of DNNs \cite{ghosh2017robust}.    
	Actually, \textit{\textbf{IMAE is neither symmetric nor bounded}}. However, it is proved to be noise-robust empirically. 
\end{itemize}
\vspace{-0.2cm}
%We provide the theoretical analysis on noise-robustness of MAE and IMAE in the supplementary material. 
These analytical discussions are demonstrated in our empirical studies in Table~\ref{table:CIFAR_noise_three} and Figures~\ref{fig:ResNet2056_Noise40_CCE_MAE_SMAE_dynamics}, \ref{fig:ResNet2056_Noise80_CCE_MAE_SMAE_dynamics}.

%\subsection{Does high loss value (will/won't/usually) backprop to high gradients for parameters}

%%%%%%%%%%%%%%%%%%%%%%%%%%%%%%%%
\begin{table*}[!t]
	%%\vspace{-0.22cm}
	\caption{
		Summary of CCE, MAE and IMAE. 
		$(\mathbf{x}, y)$ is a training example. 
		For simplicity, $p_{y} = p(y|\mathbf{x})$, and $p_{j} = p(j|\mathbf{x}), j \neq y$,  
		$\sum_{j\neq y}p_j + p_y = 1$.
		Prior analysis on loss functions is based on the loss expression or $\frac{\partial L}{\partial \mathbf{p}}$. 
		Instead, we are the first to study the differences of loss functions according to $||\frac{\partial L}{\partial \mathbf{z}}||_1$. Our empirical evidences justifies its rationality. 
		Note that we have $L(p_y)=\int \frac{\partial L}{\partial p_y} d p_y, L(1)=0$, therefore $ L(p_y)=\int_{p_y}^{1} -\frac{\partial L}{\partial p_y} dp_y$. We remark \textit{\textbf{IMAE is neither symmetric nor bounded}}, which challenges the robustness theories studied in \cite{ghosh2017robust,zhang2018generalized,wang2019symmetric}.
	}
	\centering
	\vspace{-8pt}
	\setlength{\tabcolsep}{0.1pt} % Default value: 6pt
	\fontsize{6.0pt}{6.0pt}\selectfont
	\begin{tabular}{lo uu aa yy g}
		\toprule
		&  \multicolumn{1}{c}{}
		&  
		\multicolumn{2}{c}{{$\frac{\partial L}{\partial \mathbf{p}}$}}
		&
		\multicolumn{2}{c}{{$\frac{\partial p_y}{\partial \mathbf{z}}$}}
		&
		\multicolumn{2}{c}{{$\frac{\partial L}{\partial \mathbf{z}}=\sum_{j=1}^{C} \frac{\partial L}{\partial p_j} \times  \frac{\partial p_j}{\partial \mathbf{z}}$}}
		& \multicolumn{1}{c}{}
		\\
		\cmidrule(l{2pt}r{2pt}){3-4}
		\cmidrule(l{2pt}r{2pt}){5-6}
		\cmidrule(l{2pt}r{2pt}){7-8}
		% next line
		&  \multicolumn{1}{c}{
			{\multirow{-3}{*}{\makecell{Loss Expression\\ $L = L(p_y)=$\\$\int_{p_y}^{1} -\frac{\partial L}{\partial p_y} dp_y$}} }
		}
		& \multicolumn{1}{c}{$\frac{\partial L}{\partial p_y}$} 
		& \multicolumn{1}{c}{\makecell{$\frac{\partial L}{\partial p_j},$\\$j\neq y$}}  
		& \multicolumn{1}{c}{ $\frac{\partial p_y}{\partial \mathbf{z}_{y}}$}
		&\multicolumn{1}{c}{\specialcell{$\frac{\partial p_y}{\partial \mathbf{z}_{j}}$, \\$j\neq y$} }
		& \multicolumn{1}{c}{ $\frac{\partial L}{\partial \mathbf{z}_{y}}$}
		&\multicolumn{1}{c}{\specialcell{$\frac{\partial L}{\partial \mathbf{z}_{j}}$, \\$j\neq y$} }
		& \multicolumn{1}{c}{\multirow{-3}{*}{\makecell{$||\frac{\partial L}{\partial \mathbf{z}}||_1$}}} \\
		%next line
		\midrule
		\multirow{1}{*}{CCE}
		& $-\log p_y $  & $-\frac{1}{p_y}$ & 0 & $p_y(1-p_y)\text{~~}$ & $-p_yp_j$ & $p_y-1$ & $p_j$ & $2
		%\times
		(1-p_y)$
		\\
		\multirow{1}{*}{MAE}
		& $2(1-p_y)$ & -2 & 0   & $p_y(1-p_y)\text{~~}$ & $-p_yp_j$ 
		& $2 p_y
		(p_y-1)$& $2 p_y
		p_j$ &
		$4
		%\times
		p_y 
		%\times
		(1-p_y)$
		\\
		%\midrule
		\multirow{1}{*}{IMAE}
		& \makecell{  $\int_{p_y}^{1} \frac{
				%w_\mathrm{IMAE} (\mathbf{x})
				\exp(T   
				%\times
				p_y 
				%\times
				(1-p_y))
			}{2 p_y
				(1-p_y)} dp_y$} & $\frac{
			%w_\mathrm{IMAE} (\mathbf{x})
			\exp(T   
			%\times
			p_y 
			%\times
			(1-p_y))
		}{2 p_y
			(p_y-1)}$ & 0   & $p_y(1-p_y)\text{~~}$ & $-p_yp_j$
		
		& $\frac{
			%w_\mathrm{IMAE} (\mathbf{x})
			\exp(T   
			%\times
			p_y 
			%\times
			(1-p_y))
		}{-2} $ 
		& 
		$\frac{
			%w_\mathrm{IMAE} (\mathbf{x})
			\exp(T   
			%\times
			p_y 
			%\times
			(1-p_y))
			p_j}{2(1-p_y)}$
		
		&\makecell{$
			%w_\mathrm{IMAE} (\mathbf{x}) = 
			$\\$ \exp(T   
			%\times
			p_y 
			%\times
			(1-p_y))$}
		\\
		
		\bottomrule
	\end{tabular}
	\label{table:GR_differentiation}
	\vspace{-0.1cm}
\end{table*}
%%%%%%%%%%%%%%%%%%%%%%%%%%%%%%%%%\\

%start-------------------------------------------------------------------------
\section{Experiments}
\label{sec:experiments}

We demonstrate the effectiveness of IMAE as follows: 
%image classification on CIFAR-10 \cite{krizhevsky2009learning} and video-based person re-identification on MARS \cite{zheng2016mars}: 

\noindent
\textbf{Outperforming the state-of-the-art.} 
IMAE is compared with recent baselines in Sections~\ref{sec:cifar_100_synthetic_SOTA} and ~\ref{sec:clothing1M_SOTA} in different scenarios: (1) Clean labels; (2) Synthetic symmetric and asymmetric noisy labels; (3) Realistic agnostic noise. 

\noindent
\textbf{Analysis of the training dynamics of IMAE against CCE and MAE.} 
We thoroughly visualise and compare the training dynamics of IMAE, CCE and MAE in Section~\ref{sec:cifar10_experiments} for empirical justification.

\textbf{Supplementary studies.} 
In our supplementary material, we further prove IMAE's effectiveness by: (1) The results on a \textit{video retrieval task} (video person re-identification); (2) The results of \textit{different stochastic optimisers}; (3) 
\textit{The ablation study of} $T$.

\subsection{Image Classification on CIFAR-100 with Synthetic Noise}
\label{sec:cifar_100_synthetic_SOTA}

\textbf{Dataset.} CIFAR-100 \cite{krizhevsky2009learning} contains 100 classes,  500 images per class for training and 100 images per class for testing. The image size is $32\times 32$.

\textbf{Synthetic label noise generation.} 
(1) Class-independent (uniform or symmetric) noise: With a probability of $r$, the label of each image is replaced by  one of the other class labels uniformly. 
(2) Class-dependent (non-uniform or asymmetric) noise:
The 100 classes of CIFAR-100 are grouped into 20 coarse ones. Every coarse one has 5 fine classes.
Following \cite{wang2019symmetric}, we first randomly select 2 out of 5 classes, and then their labels are flipped to each other with a probability of $r$. 
%An image's label is uniformly flipped to one of the other four fine classes of the same coarse class with a probability $r$. 
%
$r$ denotes the noise rate. 
All instances generated from the same original image by data augmentation share the same label.  
All test labels are kept intact.  

\textbf{Implementation details.} 
We follow the settings of recent SL \cite{wang2019symmetric} and train ResNet44 \cite{he2016deep} for a fair comparison with their reported results.
We also use the same data augmentation techniques: random horizontal flips and crops of $32\times 32$ on the images after being padded with 4 pixels on each side.
All networks are trained using SGD with a momentum of 0.9, a weight decay of 0.0005 and an initial learning rate of 0.1.

\textbf{Baselines.} IMAE is compared against standard CCE, MAE, and six recent robust training baselines: 
1) Forward (or Backward) applies a noise-transition matrix to multiply the network's predictions (or losses) for label correction purpose \citep{patrini2017making}; 
2) Bootstrapping learns on new labels generated by a convex combination (soft or hard combinations) of the original ones and their predictions \citep{reed2015training}.  
3) D2L achieves noise-robustness by restricting the dimensionality expansion of learned subspaces during training \citep{ma2018dimensionality};  
4) SL boosts CCE with a noise-robust counterpart, i.e., reverse cross entropy \citep{wang2019symmetric};
5) GCE aims to achieve a balance between MAE and CCE \cite{zhang2018generalized};
6) Label Smoothing (LS) trains DNNs on softly smoothed labels instead of one-hot ones;  
We remark that \cite{lee2019robust} is not benchmarked for two reasons: (1) The used network is not ResNet-44 by checking with the authors; (2) The proposed algorithm is orthogonal to ours because it targets at the inference stage and is a generative classifier on top of pre-trained deep representations. Our IMAE focuses on the training stage and is a softmax-based neural classifier. 

\textbf{Results.} 
We display the results in Tables~\ref{table:cifar100_SOTA_Symmetric} and~\ref{table:cifar100_SOTA_Asymmetric}.
We observe that IMAE is superior to the state-of-the-art. We fix the random seed as 123 and do not use any random computational accelerator for the purpose of exact reproducibility.

%%%%%%%%%%%%%%%%%%%%%%%%%%%%%%%%
\begin{table}[!t]
	\caption{
		The results on CIFAR-100 using ResNet44.
		%The 1st and 2nd blocks show results from SL \cite{wang2019symmetric} and D2L \cite{ma2018dimensionality}, respectively. 
		Results from SL and D2L are different due to different optimisation details. 
		In our experiments, we fix the random seed as 123 and do not use any random computational accelerator for the purpose of exact reproducibility.  
		The best results on each block and our IMAE are bolded. %\\
		%Colored line is the most basic baseline where examples have the same derivative's magnitude when $\beta=0$.	
	}
	\label{table:cifar100_SOTA_Symmetric}
	\centering
	\vspace{-0.2cm}
	\fontsize{9.6pt}{9.6pt}\selectfont
	\setlength{\tabcolsep}{3.8pt} % Default value: 6pt
	\begin{tabular}{cccccc}
		\toprule
		&\multirow{2}{*}{Method} & \multirow{2}{*}{\specialcell{Clean\\Labels}} & \multicolumn{3}{c}{Symmetric Noisy Labels} \\
		
		\cmidrule(l{2pt}r{2pt}){4-6}
		%\cmidrule(l{2pt}r{2pt}){7-9}
		& & & $r$=0.2 & $r$=0.4 & $r$=0.6 \\
		
		\midrule

		\multirow{7}{*}{\specialcell{Results \\From \\ SL}}
		&CCE & 64.3 & 59.3 & 50.8 & 25.4 \\
		&LS & 63.7 & 58.8 & 50.1 & 24.7 \\
		&Boot-hard & 63.3 & 57.9 & 48.2 & 12.3 \\
		&Forward & 64.0 & 59.8 & 53.1 & 24.7 \\
		&D2L & 64.6 & 59.2 & 52.0 & 35.3 \\
		&GCE & 64.4 & 59.1 & 53.3 & 36.2 \\
		&SL & \textbf{66.8} & \textbf{60.0} &  \textbf{53.7} &  \textbf{41.5} \\
		
		\cmidrule{2-6}
		\multirow{7}{*}{\specialcell{Results \\From \\ D2L}}
		&CCE & {68.2} & {52.9} & {42.9} & {30.1} \\
		&Boot-hard & {68.3} & {58.5} & {44.4} & {36.7} \\
		&Boot-soft & {67.9} & {57.3} & {41.9} & {32.3} \\
		&Forward & {68.5} & {60.3} & {51.3} & {41.2} \\
		&Backward & {68.5} & {58.7} & {45.4} & {34.5} \\
		&D2L & \textbf{68.6} & \textbf{62.2} & \textbf{52.0} & \textbf{42.3} \\
		
		\midrule
		%\cmidrule{1-5}
		\multirow{3}{*}{\specialcell{Our \\Trained \\Results}}
		&CCE & \textbf{70.0} & {60.4} & {53.2} & {42.1} \\
		&MAE & {8.2} & {6.4} & {7.3} & {5.2} \\
		
		&IMAE & \textbf{69.2} & \textbf{63.4} & \textbf{54.7} & \textbf{43.9} \\
		
		%DM($\lambda=2$) & {68.3} & {60.0} & {51.3} & {43.7} \\

		\bottomrule
	\end{tabular}
	%\vspace{-0.3cm}
\end{table}
%%%%%%%%%%%%%%%%%%%%%%%%%%%%%%%%%

%%%%%%%%%%%%%%%%%%%%%%%%%%%%%%%%
\begin{table}[!t]
	\caption{
		The results on CIFAR-100 using ResNet44.
		The best results on each block are bolded. 
	}
	\label{table:cifar100_SOTA_Asymmetric}
	\centering
	\vspace{-0.2cm}
	\setlength{\tabcolsep}{1.8pt} % Default value: 6pt
	\fontsize{9.6pt}{9.6pt}\selectfont
	\begin{tabular}{ccccc}
		\toprule
		\multirow{2}{*}{} & \multirow{2}{*}{Method} & \multicolumn{3}{c}{Asymmetric Noisy Labels} \\
		
		\cmidrule{3-5}
		& & $r$=0.2 & $r$=0.3 & $r$=0.4 \\
		\midrule
		
		\multirow{7}{*}{\specialcell{Results From \\ SL \\ \cite{wang2019symmetric}}}
		&CCE & 63.0 & 63.1 & 61.9 \\
		&LS & 63.0 & 62.3 & 61.6 \\
		&Bootstrap & 63.4 & 63.2 & 62.1 \\
		&Forward & 64.1 & 64.0 & 60.9  \\
		&D2L & 62.4 & 63.2 & 61.4 \\
		&GCE & 63.0 & 63.2 & 61.7 \\
		&SL & \textbf{65.6} &  \textbf{65.1} & \textbf{63.1} \\
		
		\midrule
		\multirow{3}{*}{\specialcell{Our trained \\Results}}
		&CCE & {66.4} & {64.7} & {60.3}\\
		&MAE & {7.3} & {6.3} & {7.3}\\
		&IMAE & \textbf{67.5} & \textbf{65.8} & \textbf{63.3}\\
		\bottomrule
	\end{tabular}
\end{table}
%%%%%%%%%%%%%%%%%%%%%%%%%%%%%%%%%

\begin{table*}[!t]
	\caption{
		%\small
		Classification accuracy (\%) on Clothing1M with ResNet50 \cite{he2016deep}. 
		%We follow exactly the same setup as the compared methods.
		The leftmost block's results are from SL \cite{wang2019symmetric} while the middle block's are from Masking \cite{han2018masking}.
	}
	\centering
	%\vspace{-0.2cm}
	\setlength{\tabcolsep}{5pt} % Default value: 6pt
	\fontsize{7.8pt}{7.8pt}\selectfont
	\begin{tabular}{cccccc|cc|cccc}
		\toprule
		
		\multirow{2}{*}{{CCE }} & \multirow{2}{*}{\makecell{Boot-hard}} & \multirow{2}{*}{{Forward} } & 
		\multirow{2}{*}{{D2L}} & 
		\multirow{2}{*}{{ GCE } }  & 
		\multirow{2}{*}{{SL}} & 
		\multirow{2}{*}{\makecell{S-adaptation}} &
		\multirow{2}{*}{ Masking} &
		\multirow{2}{*}{ \makecell{Joint \\Optim.}} & 
		\multicolumn{3}{c}{Our trained results}  
		\\
		\cmidrule{10-12}
		
		&  &  
		&  &   &  &  &  &  & CCE  & MAE & IMAE\\
		\midrule
		68.8 & 68.9 & 69.8& 69.5 & 69.8 & 71.0  
		& 70.3 & 71.1 & 72.2 & 71.7 &  39.7 & \textbf{73.2} \\
		\bottomrule
	\end{tabular}
	\label{table:Clothing1M_competitors}
	%\vspace{-0.3cm}
\end{table*}

\subsection{Image Classification on Clothing1M with Realistic Unknown Noise}
\label{sec:clothing1M_SOTA}
\noindent
\textbf{Dataset.} Clothing1M \cite{xiao2015learning} contains one million clothing images of fourteen classes from online shopping websites. \textit{Its noise type is agnostic.} 
%About 61.54\% training labels are correct, i.e., 
The noise rate is around 38.46\%.
%\cite{xiao2015learning}. 
Additionally, it includes 50k, 14k, and 10k images with clean labels for training, validation, and testing, respectively. To compare fairly with existing algorithms without exploiting auxiliary information from trusted clean data, we also train only on the noisy training data.

\noindent
\textbf{Implementation details.}
%We follow exactly the same settings as \cite{patrini2017making,tanaka2018joint}.
%and compare with their reported results. 
%Concretely, we 
We follow \cite{patrini2017making,tanaka2018joint,wang2019symmetric} and train ResNet50 initialised by pretrained ImageNet model \cite{russakovsky2015imagenet}. We apply an SGD optimiser with a momentum of 0.9 and a weight decay of 0.00002. We set the initial learning rate to 0.01 and divide it by 10 after 10k and 15k iterations. We stop training at 30k iterations. Regarding data augmentation, a raw input image is warped to 256$\times$256, followed by a random crop of 227$\times$227 and a random horizontal mirroring. The batch size is 84. 
Every program is run on a single Tesla V100 GPU with 32 GB RAM.

\noindent
\textbf{Competitors.} Some recent baselines are compared: %under similar settings: 
%1) Standard CCE;
%\cite{patrini2017making} 
%We also train it ourselves for reference; 
%2) Bootstrapping Soft learns on new labels generated by a soft convex combination of the original labels and their predicted ones \cite{reed2015training}; 
%3) Forward multiplies a network's predictions with a noise-transition matrix for loss correction purpose \cite{patrini2017making}; 
%4) Bilevel Optimisation minimises empirical errors on a training set and a testing set simultaneously \cite{jenni2018deep};
1) S-adaptation explicitly estimates latent true labels by an additional softmax layer \cite{goldberger2017training};
2) Masking speculates the structure of a noise-transition matrix with human cognition \cite{han2018masking}; 
3) Joint Optim. iteratively optimises model's parameters and latent true labels \cite{tanaka2018joint}.
%, where two regularisation terms are adjusted in practice 
Others are introduced in Section~\ref{sec:cifar_100_synthetic_SOTA}.
Note that \cite{han2019deep} corrects labels gradually and \cite{li2019learning} exploits meta-learning. They are not technically related and not benchmarked consequently. 

\noindent
\textbf{Results.} 
%We compare with the state-of-the-art on Clothing1M 
We display the results in Table~\ref{table:Clothing1M_competitors}.
IMAE outperforms the state-of-the-art, which proves IMAE's effectiveness under real-world scenarios with agnostic noise. 
Beyond, we remark that IMAE is much simpler than those competitors except CCE, MAE.

\subsection{Empirical Analysis of IMAE Against Basic Baselines CCE and MAE on CIFAR-10}
\label{sec:cifar10_experiments}
\noindent
\textbf{Dataset.} CIFAR-10 \cite{krizhevsky2009learning} contains 10 classes,  5k images per class for training and 1k images per class for testing. The image size is $32\times 32$.

\noindent
\textbf{Implementation details \footnote{Our purpose is to study the behaviours of CCE, MAE and IMAE  on CIFAR-10 instead of pushing its state-of-the-art results.}.}
We follow the study on CIFAR-10 in \cite{he2016deep}, which means we use exactly the same architectures (ResNet20, ResNet56) and training settings: a weight decay of 0.0001, a momentum of 0.9, a batch size of 128. The learning rate starts at 0.1, then is divided by 10 at 32k and 48k iterations. Training stops at 100k iterations. 
{Data augmentation is the same as CIFAR-100.}
For IMAE, \textit{without tuning $T$ case by case}, we fix $T=0.5$ when training data is clean and $T=8$ when noise exists although noisy rate is different.\footnote{More discussion about the hyper-parameter $T$ is given in our supplementary material.}

A well-accepted way to improve data fitting ability is increasing a model's capacity. Therefore, we train a shallower net ResNet20 and a deeper net ResNet56 
%with similar structure 
for better analysis.

\subsubsection{CIFAR-10 with intact labels}
In Table~\ref{table:CIFAR_noise_three}, we first compare IMAE with CCE and MAE on clean CIFAR-10 using different nets (ResNet20, ResNet56). 
%From shallower to deeper architectures, 
We observe that IMAE is competitive with CCE and outperforms MAE significantly.

\subsubsection{CIFAR-10 with corrupted labels}
Following \cite{zhang2017understanding,arpit2017closer}, we test the robustness of deep models against corrupted labels. 
%The memorisation of noisy labels has been studied from the perspective of regularisation \cite{zhang2017understanding,arpit2017closer} and data augmentation \cite{zhang2018mixup}. In this work, we present an analysis of it from the perspective of sample weighting in loss functions. Interestingly, IMAE alleviates both overfitting and underfitting by weighting training samples properly.
%
%\noindent
%\textbf{Corrupted training sets with symmetric noise.} 
%We corrupt CIFAR-10 training labels following \cite{zhang2017understanding,zhang2018mixup}. 
We evaluate on uniform noise because it is more challenging than asymmetric noise which is verified in \cite{vahdat2017toward}.

\noindent 
\textbf{Majority voting assumption.}
When generating uniform noise on CIFAR-10, even up to 80\% noise rate, clean examples are still the majority because 80\% labels are corrupted to other 9 classes evenly. 
\textit{We remark that the majority voting is our reasonable assumption}. We believe that if the noise becomes the majority, it is hard to discover meaningful patterns. Being natural and intuitive, the majority define the meaningful data patterns to learn.

\begin{figure*}[!t]
	\centering
	\vspace{-0.3cm}
	\begin{subfigure}[t!]{0.33\linewidth}
		\centering
		\includegraphics[width=0.780\linewidth]{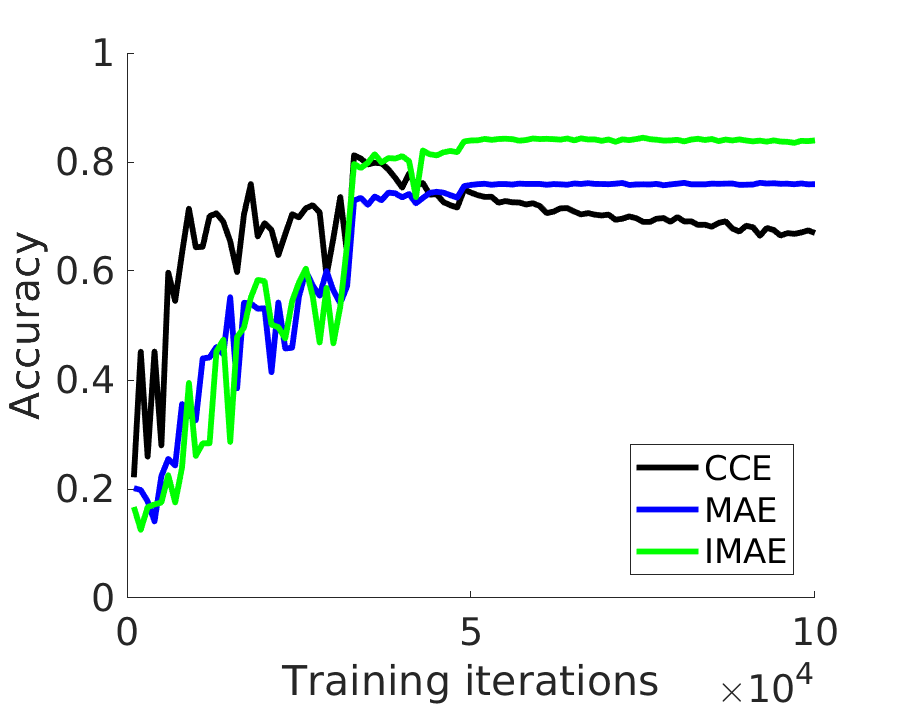}
		\caption{ResNet20: Testing set (higher is better).}
	\end{subfigure}%
	\begin{subfigure}[t!]{0.33\linewidth}
		\centering
		\includegraphics[width=0.82\linewidth]{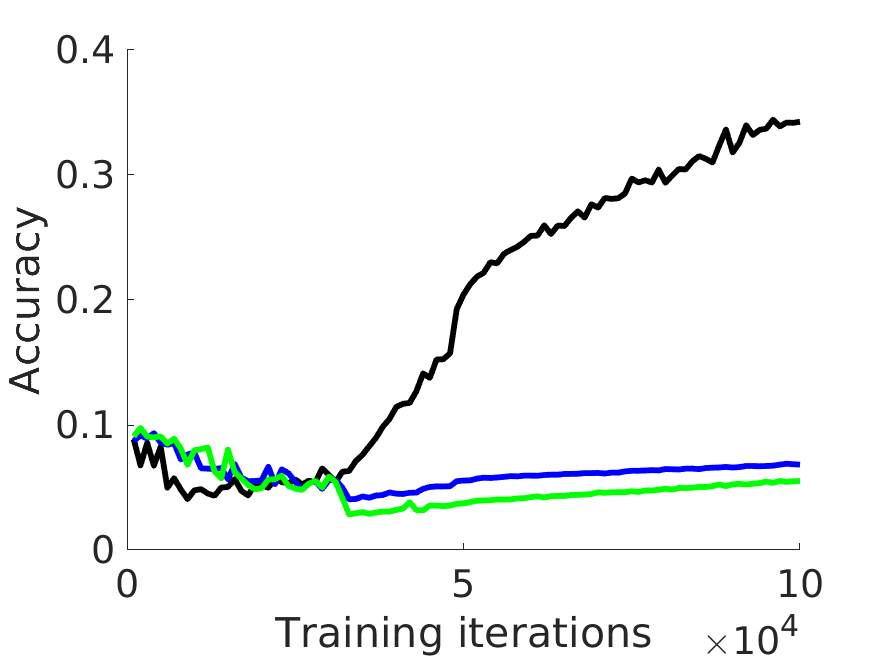}
		\caption{ResNet20: Noisy subset (lower is better).}
	\end{subfigure}
	\begin{subfigure}[t!]{0.33\linewidth}
		\centering
		\includegraphics[width=0.780\linewidth]{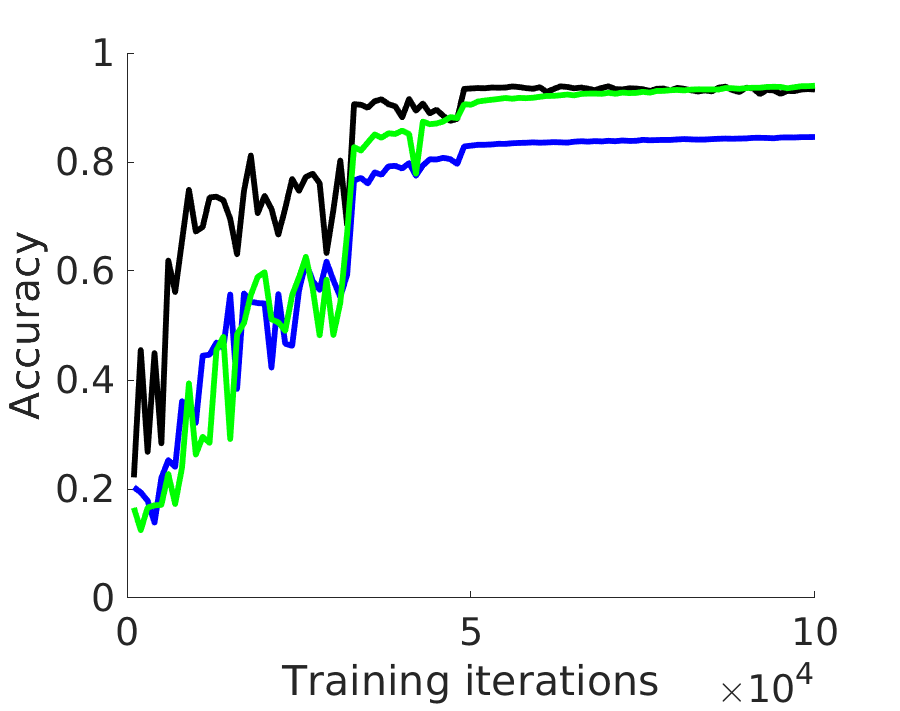}
		\caption{ResNet20: Clean subset (higher is better).}
	\end{subfigure}
	\begin{subfigure}[t!]{0.33\linewidth}
		\centering
		\includegraphics[width=0.780\linewidth]{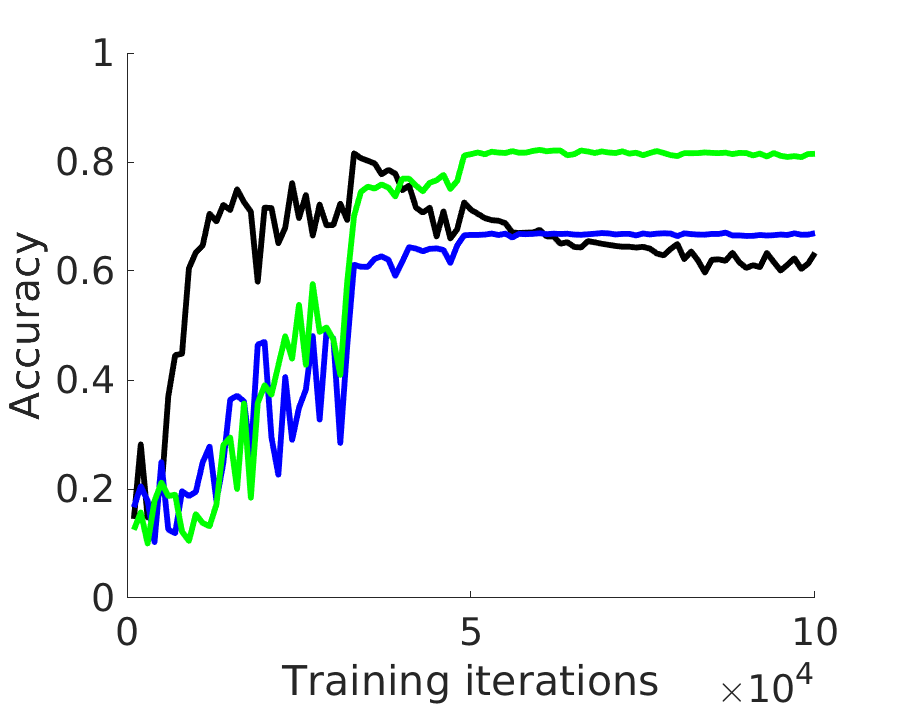}
		\caption{ResNet56: Testing set (higher is better).}
	\end{subfigure}%
	\begin{subfigure}[t!]{0.33\linewidth}
		\centering
		\includegraphics[width=0.825\linewidth]{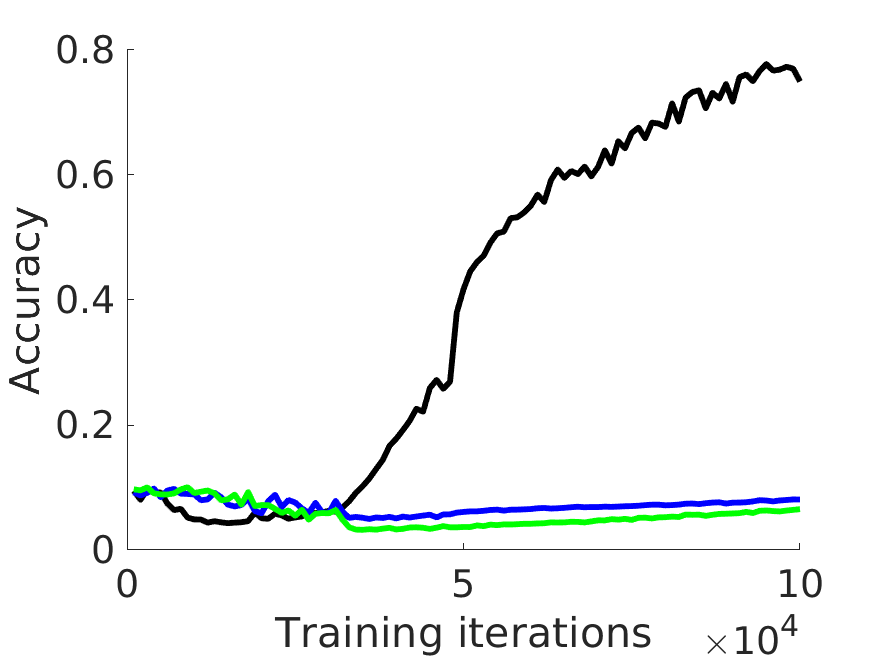}
		\caption{ResNet56: Noisy subset  (lower is better).}
	\end{subfigure}
	\begin{subfigure}[t!]{0.33\linewidth}
		\centering
		\includegraphics[width=0.780\linewidth]{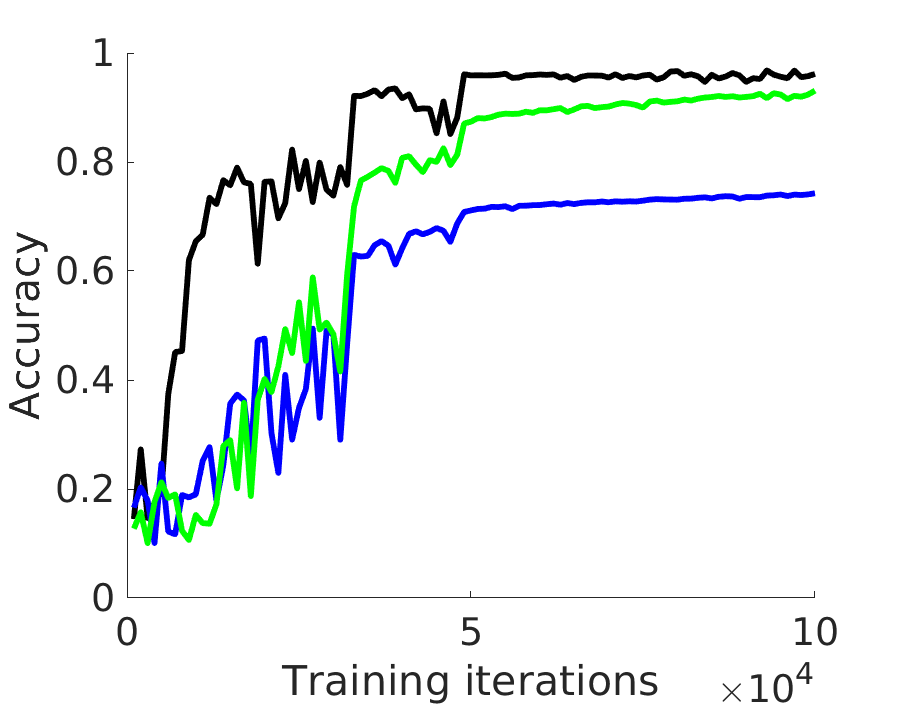}
		\caption{ResNet56: Clean subset (higher is better).}
	\end{subfigure}
	\caption{CIFAR-10 with noise rate $r=40\%$. The accuracies on testing set, noisy subset and clean subset of training set along with training iterations.
		The legend on the top left is shared by all subfigures. 
		\textit{Better viewed in colour.} }
	\label{fig:ResNet2056_Noise40_CCE_MAE_SMAE_dynamics}
	\vspace{-0.1cm}
\end{figure*}

\noindent 
\textbf{Results.} 
The results are summarised in Table~\ref{table:CIFAR_noise_three}. 
For more comprehensive and clear comparison, we display the training dynamics in Figures~\ref{fig:ResNet2056_Noise40_CCE_MAE_SMAE_dynamics} (40\% noise) and ~\ref{fig:ResNet2056_Noise80_CCE_MAE_SMAE_dynamics} (80\% noise) of the supplementary material. 
\textit{Note that general learning objectives are high final testing accuracy, low accuracy on the noisy training subset, and high accuracy on the clean training subset.} 
Therefore, we report the hybrid accuracy on the combination of testing set and clean training set. 
%Furthermore, it considers not only fitting the noisy subset less, but also predicting those abnormal examples in the noisy subset correctly. 
We have the following observations:
\begin{itemize}
	\vspace{-0.2cm}
	\item Regarding CCE's test accuracies, the best is always much higher than the final. 
	%according to   Table~\ref{table:CIFAR_noise_three} and Figures \ref{fig:ResNet2056_Noise40_CCE_MAE_SMAE_dynamics}, \ref{fig:ResNet2056_Noise80_CCE_MAE_SMAE_dynamics}. 
	%On the one hand, 
	%When fixing the architecture, their gap becomes larger as the noise rate $r$ increases from 20\% to 80\%. 
	%On the other hand, 
	%While fixing the noise rate, the gap also becomes more dramatic when network is deeper. 
	In Figures \ref{fig:ResNet2056_Noise40_CCE_MAE_SMAE_dynamics} and \ref{fig:ResNet2056_Noise80_CCE_MAE_SMAE_dynamics}, \textit{as training goes, CCE always tries to fit the noisy training subset better.} Therefore, CCE learns a lot of error information  when severe noise exists.  
	When it comes to 
	MAE and IMAE, the gap between the best and final accuracies is significantly smaller than that of CCE regardless of net's capacity. 
	
	\vspace{-0.2cm}
	\item The training accuracies on both noisy and clean subsets are compared.
	%in Table~\ref{table:CIFAR_noise_three} and Figures~\ref{fig:ResNet2056_Noise40_CCE_MAE_SMAE_dynamics},~\ref{fig:ResNet2056_Noise80_CCE_MAE_SMAE_dynamics}. 
	%
	Whatever the noise rate and net's capacity are, CCE fits the noisy subset much more. 
	%Moreover, CCE is inclined to fit it better as training goes. Accordingly, its testing accuracy drops gradually. 
	%
	Although MAE fits the noisy subset much less, it fits the clean subset worst. 
	Instead, our IMAE fits the noisy subset little and the clean subset competitively with CCE.
	
	\vspace{-0.2cm}
	\item IMAE obtains the best hybrid accuracy consistently. 
\end{itemize}
\vspace{-0.2cm}

%These observations consolidate our analytical interpretation of MAE's underfitting and IMAE's effectiveness in practice.

%%%%%%%%%%%%%%%%%%%%%%%%%%%%%%%%
\begin{table*}[!t]
	\caption{
		%\textbf{To be revised: different noisy data, 40\%, 80\%, 0, Different column (analysis);Diffferent dynamics. }
		%\\
		Results (\%) of CCE, MAE and IMAE on CIFAR-10 with different noise rates. For classification accuracy on the testing set, we show the best result achieved during training and the final result when training stops, which are indicated by `Best' and `Final', respectively. For training accuracy, the results on noisy and clean subsets are displayed. 
		The hybrid accuracy represents the result on the combination of testing set and clean training set. 
		We report training and hybrid accuracies of the final model when training terminates. 
		\textit{The ultimate objective is to achieve high hybrid accuracy, since both training and testing data points may occur in a deployed system. }
		%Since the overlap rate between corrupted and intact sets is $(1-r)$, the intact training set can be regarded as a validation set.
		The best result on each column block is bolded. 
		`--' indicates there is no noisy subset.
	}
	\label{table:CIFAR_noise_three}
	\centering
	\vspace{-0.2cm}
	\fontsize{7.8pt}{7.8pt}\selectfont
	\setlength{\tabcolsep}{8pt} % Default value: 6pt
	\begin{tabular}{lccccccc}
		\toprule
		\multirow{2}{*}{Backbone}& \multirow{2}{*}{Noise rate} & \multirow{2}{*}{Loss} & \multicolumn{2}{c}{Testing accuracy} & \multicolumn{2}{c}{\makecell{Training accuracy: Naive fitting 
				%\\(Final model)
		} } & \multirow{2}{*}{\makecell{Hybrid accuracy: \\Meaningful patterns}}\\
		
		\cmidrule(l{2pt}r{2pt}){4-5}
		\cmidrule(l{2pt}r{2pt}){6-7}
		& & & Best & Final & Noisy subset & Clean subset \\
		\midrule
		
		\multirow{9}{*}{ResNet20}
		&\multirow{3}{*}{0\%}
		&CCE & 91.5 & 91.3 & -- & \textbf{100} & \textbf{98.5}\\
		&&MAE & 89.3 & 89.2 & -- & 95.8 & 94.7\\
		%&&IMAE-4 & 87.3 & 85.8 & -- & -- &  T=4\\
		&&IMAE & \textbf{91.7} & \textbf{91.4} & -- & 99.8 & 98.4\\
		%&&IMAE-16 & 87.9 & 87.9 & -- & -- &  T=16\\
		\cmidrule{2-8}
		&\multirow{3}{*}{40\%}
		&CCE & 81.2 & 67.0 & 34.3 & 93.3 & 72.6\\
		&&MAE & 76.2 & 75.9 & 6.8 & 84.6 & 79.7\\
		%&&IMAE-4 & 80.4 & 70.7 &  & -- &  T=4\\
		&&IMAE & \textbf{84.3} & \textbf{84.0} & \textbf{5.5} & \textbf{94.0} & \textbf{88.2} \\
		%&&IMAE-16 & 82.3 & 81.8 & -- & -- &  T=16\\
		\cmidrule{2-8}
		&\multirow{3}{*}{80\%}
		&CCE & 43.0 & 20.3 & 38.3 & 57.0 & 22.0\\
		&&MAE & 27.7 & 27.5 & \textbf{9.7} & 29.4 & 27.8 \\
		%&&IMAE-4 & 64.6 & 47.7 & -- & -- &  T=4\\
		&&IMAE & \textbf{52.0} & \textbf{41.0} & 16.8 & \textbf{64.8} & \textbf{41.5}\\
		%&&IMAE-16 & 15.8 & 15.8 & -- & -- &  T=16\\
		
		\midrule

		\multirow{9}{*}{ResNet56}
		&\multirow{3}{*}{0\%}
		&CCE & \textbf{92.4} & \textbf{92.2} & -- & \textbf{100} & \textbf{98.7}\\
		&&MAE & 89.0 & 89.0 & -- & 96.1 & 94.9\\
		%&&IMAE-4 & 87.3 & 67.5 & -- & -- &  T=4\\
		&&IMAE & 92.2 & \textbf{92.2} & -- & 99.8 & 98.5\\
		%&&IMAE-16 & -- & -- & -- & -- &  T=16\\
		\cmidrule{2-8}
		&\multirow{3}{*}{40\%}
		&CCE & 81.6 & 63.3 & 75.0 & \textbf{96.2} & 63.6\\
		&&MAE & 67.0 & 66.9 & 8.1 & 74.3 & 70.2\\
		%&&IMAE-4 & 80.4 & 70.7 & -- & -- &  T=4\\
		&&IMAE & \textbf{82.2} & \textbf{81.5} & \textbf{6.5} & 93.1  & \textbf{86.5}\\
		%&&IMAE-16 & 83.4 & 82.9 & -- & -- &  T=16\\
		\cmidrule{2-8}
		
		&\multirow{3}{*}{80\%}
		&CCE & \textbf{38.2} & 16.4 & 52.5 & \textbf{62.3} & 18.7\\
		&&MAE & 15.2 & 15.1 & \textbf{9.6} & 15.6 & 15.1 \\
		%&&IMAE-4 & 66.7 & 48.6 & -- & -- &  T=4\\
		&&IMAE & 37.1 & \textbf{34.0} & 13.0 & 44.7 & \textbf{34.8} \\
		%&&IMAE-16 & -- & -- & -- & -- &  T=16\\
		\bottomrule
	\end{tabular}
\end{table*}
%%%%%%%%%%%%%%%%%%%%%%%%%%%%%%%%%

%%%%%%%%%%%%ResNet2056_Noise40_CCE_MAE_SMAE_dynamics%%%%%%%%%%%%%%%%%%%%%%%%%%%%
%%%%%%%%%%%ResNet2056_Noise40_CCE_MAE_SMAE_dynamics%%%%%%%%%%

\section{Related Work}

IMAE is a family of robust loss functions, inspired by the intrinsic example weighting scheme of MAE. 
Therefore, 
our work is related to some prior work about example weighting  and robust loss functions.
\subsection{Example Weighting} 
In \cite{ren2018learning}, a meta-learning algorithm weights data points according to their gradient directions. The meta-learning algorithm is optimised on a clean validation set. In contrast, our IMAE assigns weights to samples based on their gradient magnitude and does not require extra clean set. 
MentorNet \cite{jiang2018mentornet} learns data-driven weighting scheme, which guides StudentNet to focus on samples whose labels are more trustful. 
%Instead of training an extra network like MentorNet \cite{jiang2018mentornet}, we focus on the loss function, which is easier in practice. 
In Active Bias \cite{chang2017active} and Focal Loss \cite{lin2017focal}, uncertain and hard examples are emphasised, respectively. 
%Therefore, their weighting schemes are considerably different. 
Other related work on weighting samples includes curriculum learning \cite{bengio2009curriculum}, self-paced learning \cite{kumar2010self}, and hard examples mining \cite{shrivastava2016training,wang2019deep}. 
\textit{In summary, what makes ours special is that the weighting scheme inherits from MAE, which is naturally built-in in the loss function without intuitive designing.}

\subsection{Noise-Robust Theorems on Loss Functions} 
Noise-robust theorems on loss functions from the angle of symmetric and bounded conditions on loss values have been studied recently \cite{ghosh2017robust,zhang2018generalized,wang2019symmetric}. 
Does a robust loss function have to be symmetric or bounded? 
The answer is NO according to this work. 
Although IMAE is neither symmetric nor bounded, we have extensive empirical studies to support its effectiveness. 

%While GCE analyses the gradient magnitude w.r.t. probabilities and tries to achieve a balance between CCE and MAE. 
%However, GCE requires data pruning and alternative convex search,  %\cite{bazaraa2013nonlinear},
%making it complex in practice. %We implement the loss proposed in \cite{zhang2018generalized} as shown in Table~\ref{table:MARS_ReID} for reference. 
%Contrastively, we study the gradient magnitude w.r.t. logit vector, which indicates one example's weight. 
%Interestingly, it leads to new conclusions, which motivate us to propose IMAE, a much simpler solution.  
%The empirical studies justify our viewpoint is more effective and reasonable. 

\vspace{-0.2cm}
\section{Conclusion}
\vspace{-0.2cm}
%It is imperative to robustly learn meaningful patterns when arbitrary abnormal training data points exist. 
%from the angle of built-in sample weighting in loss functions. 
%
%We present a fundamental analysis that CCE inherently focuses on low-probability examples while MAE prioritises medium-probability ones. However, MAE's differentiation degree of samples is too small, leading to informative ones cannot contribute enough against non-informative ones. Consequently, CCE easily overfits training data while MAE generally underfits it. 
%\\ 
%Therefore, we propose IMAE, which inherits noise-robustness from MAE and addresses its underfitting issue well. 
%
%Apart from being fundamental, extensive experiments demonstrate IMAE's effectiveness and robustness. 
%Furthermore, it is simple in practice since only loss function is modified. 
%
%
%Our first contribution is that 
%In this work, 
We firstly present a thorough study of CCE and MAE technically and empirically. Compared with previous work, we introduce our observations and new conclusions: 
1) MAE underfits to meaningful patterns; 
2) MAE is noise-tolerant because of emphasising on medium-probability (uncertain) examples instead of treating all samples equally.
Secondly, we claim gradient magnitude's variance matters. As a consequence, we propose an effective and simple solution for addressing MAE's underfitting issue while preserving its noise-robustness. 
IMAE is a family of robust loss functions whose gradient magnitude's variance is adjustable. 
%Finally, the effectiveness of IMAE is demonstrated in different cases: (1) Training data is clean; (2) Synthetic noisy labels exist; (3) Real-world agnostic noise exists. 

We remark that our empirically demonstrated claim--``Gradient Magnitude's Variance Matters''--can be applied for other algorithms as well, for example, CCE.  
However, it is beyond the scope of this work since we focus on analysing MAE and how to improve MAE here. 
We will investigate this claim in other loss functions in our future work.   

Furthermore, we have a research plan to study the effectiveness of IMAE's variants for the robustness against adversarial perturbations \cite{kurakin2017adversarial, moosavi2017universal}, e.g., incorporating IMAE's variants with iterative trimmed loss minimisation \cite{shen2019learning}. 

%In the appendices, we present complementary technical discussions, related work, and experiments.  

%start-------------------------------------------------------------------------

%end-------------------------------------------------------------------------

%{\small
%	\bibliographystyle{aaai}
%	\bibliography{ICCV2019_IMAE_DML}
%}

{\small

\bibliographystyle{iclr2021_conference}
%\medskip

\bibliography{NIPS2019_USW}
}

\newpage
\newpage
\onecolumn
\appendix

%\newpage
%{~}
%\newpage 
%\clearpage
%\title{ Supplementary Material for IMAE
%\vspace{-0.5cm}
%}
\begin{center}
	\LARGE
	{Supplementary Material for IMAE}\\
	{~}
	\\
\end{center}

%\author{}
%\maketitle
%\pagestyle{plain}
%\setcounter{page}{1}
%\setcounter{equation}{16}
%\setcounter{table}{3}
%\setcounter{figure}{4}
%\setcounter{section}{0}

%start-------------------------------------------------------------------------

%\section{IMAE's design motivation and analytical discussions}
%\vspace{-0.2cm}
%

%\section{Differentiation degree over training examples of IMAE with different $T$}
\section{The Impact of $T$ on Gradient Magnitude's Variance}

%The differentiation degree over training samples is indicated by the weighting variance. It reflects the relative contribution of one example versus another during training.   
%We show the weighting curves of IMAE with different $T$ in Figure~\ref{fig:absolute_weight_IMAE_Ts}.  

\textit{Assuming samples' probabilities are uniformly distributed}, we calculate the variances of IMAE's weighting curves with different $T$. As illustrated in Sec. 4 of the main paper, we rewrite the Eq. (\ref{eq:weight_IMAE}) (We use $e$ to replace $\exp$ for brevity):

\begin{equation}
w_{\mathrm{IMAE}} (p) =  e^{T  \cdot
	%\times
	p 
	%\times
	(1-p)},
\end{equation}
where $p$ is the probability of one randomly sampled example being predicted to its annotated label.  
According to Eq. (\ref{eq:variance_IMAE}) in the main paper, we have,

%\begin{equation}
%\begin{aligned} \displaystyle\int^{1}_{0} \mathrm{e}^{8x{\cdot}\left(1-x\right)}\, \mathrm d x = {{e^2\,\sqrt{\pi}\,\mathrm{erf}\left(\sqrt{2}\right)}\over{2^{{{3
%				}\over{2}}}}} ~=~ 4.4197196\end{aligned}
%\end{equation}

\begin{equation}
\begin{aligned}
\sigma_{\mathrm{IMAE}}&=
\displaystyle\int^{1}_{0} 
w_{\mathrm{\mathrm{IMAE}}}^{2} (p)
\, \mathrm d p
- (\displaystyle\int^{1}_{0} 
w_{\mathrm{IMAE}} (p)
\, \mathrm d p)^2 \\
&=
\displaystyle\int^{1}_{0} \mathrm{e}^{2Tp{}\left(1-p\right)}\, \mathrm d p
-
(\displaystyle\int^{1}_{0} \mathrm{e}^{Tp{}\left(1-p\right)}\, \mathrm d p)^2 \\
&=
{{\sqrt{\pi}\,\mathrm{erf}\left({{\sqrt{2T}}\over{{2}}}\right)
\,e^{{{T}\over{2}}}}\over{\sqrt{2T}}}
-
{{{\pi}\,\mathrm{erf^2}\left({{\sqrt{T}}\over{2}}\right)\,e^{{{T
}\over{2}}}}\over{{T}}}
.
\end{aligned}
\end{equation}
%\begin{equation}
%\begin{aligned} \displaystyle\int^{1}_{0} \mathrm{e}^{tx{\cdot}\left(1-x\right)}\, \mathrm d x = {{\sqrt{\pi}\,\mathrm{erf}\left({{\sqrt{t}}\over{2}}\right)\,e^{{{t
%				}\over{4}}}}\over{\sqrt{t}}} \end{aligned}
%\end{equation}
$\mathrm{erf}$ is the error function. 
Therefore we obtain the weighting variances $\sigma_{\mathrm{IMAE}}$ of IMAE with different $T$, as displayed in Table~\ref{table:variance_IMAE_T}.  
%In the main paper, in cases where corrupted training labels exist, we simply fix $T=8$ (IMAE-8) without any tuning. 
%Based our empirical studies, its weighting variance is acceptable although it is not optimal.  

\begin{table}[!h]
\vspace{-0.0cm}
\caption{
The weight variance (gradient magnitude's variance) of IMAE when $T$ changes.
}
\centering
\setlength{\tabcolsep}{3.5pt} % Default value: 6pt
\begin{tabular}{lccccccc}
\hline
$T$  & 16 & 8 & 4 & 2 & 1 & 0.5 & 0\\
\hline
$\sigma_{\mathrm{IMAE}}$ & 354.113 & 4.546 & 0.299 & 0.040 & 0.007 & 0.002 & 0\\
\hline
\end{tabular}
\label{table:variance_IMAE_T}
%\vspace{-0.4cm}
\end{table}
%%%%%%%%%%%%%%%%%%%%%%%%%%%%%%%%%

%%%%%%%%%%%%%%%%%%%%%%%%%%%%%%%%%%%%%%%%%%%%%%%%%%%%%
\begin{figure}[!h]
\centering
\includegraphics[width=0.6\linewidth]{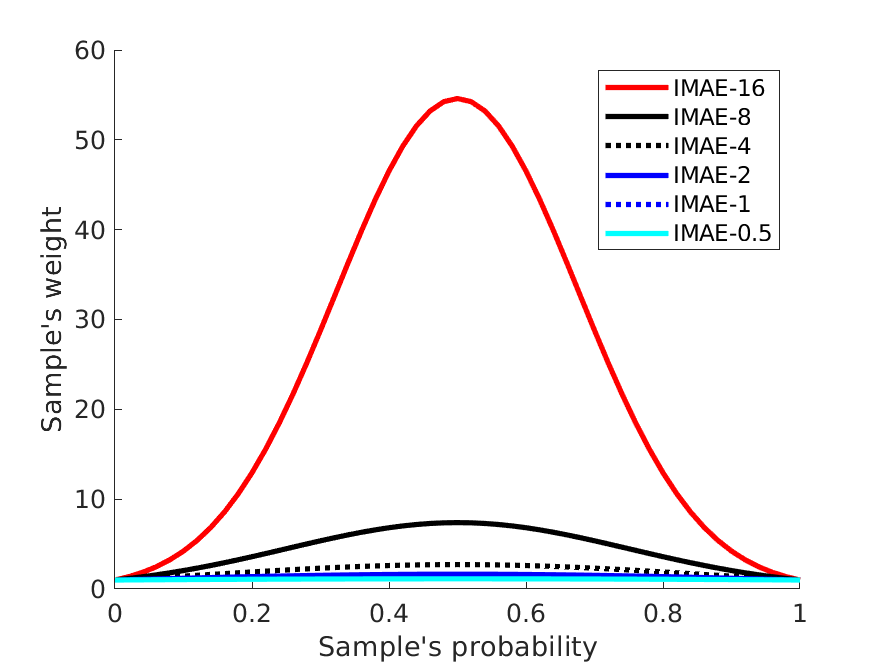}
\caption{
Sample's weight along with probability in IMAE with different $T$ (IMAE-$T$). 
%
%However, the differentiation degree over data points is too small in MAE because of its variance of weighting curve is only 0.09.
%
%Therefore, MAE cannot fit training data properly. The weight variance of IMAE-8's curve is 4.55 so that IMAE-8 differentiates samples well and addresses MAE's underfitting issue.		
The hyper-parameter $T$ controls gradient magnitude's variance, and impact ratio between examples consequently.
%IMAE own good robustness by emphasizing on medium-probability samples. 
\textit{Better viewed in colour.}
}
\label{fig:absolute_weight_IMAE_Ts}
\vspace{-0.4cm}
\end{figure}
%%%%%%%%%%%%%%%%%%%%%%%%%%%%%%%%%%%%%%%%%%%%%%%%%%%%

\section{The Impact of $T$ on Validation Accuracy}
\label{sec:test_accuracies}

We visualise and compare the effect of $T$ on CIFAR-10 test performance.
These experiments follow exactly the same settings of the main paper. %We choose different $T$ in different experiments. 

We try two cases: (1). Training labels are intact ($r=0$); (2). Training labels are corrupted randomly with a probability of 0.4 ($r=40\%$). In both cases, the test set is kept intact for evaluation. The backbone network is ResNet20.

\subsection{CIFAR-10 with intact training labels}

\begin{figure*}[t!]
\centering
\begin{subfigure}[t!]{0.48\linewidth}
\centering
\captionsetup{width=1\linewidth}
\includegraphics[width=\linewidth]{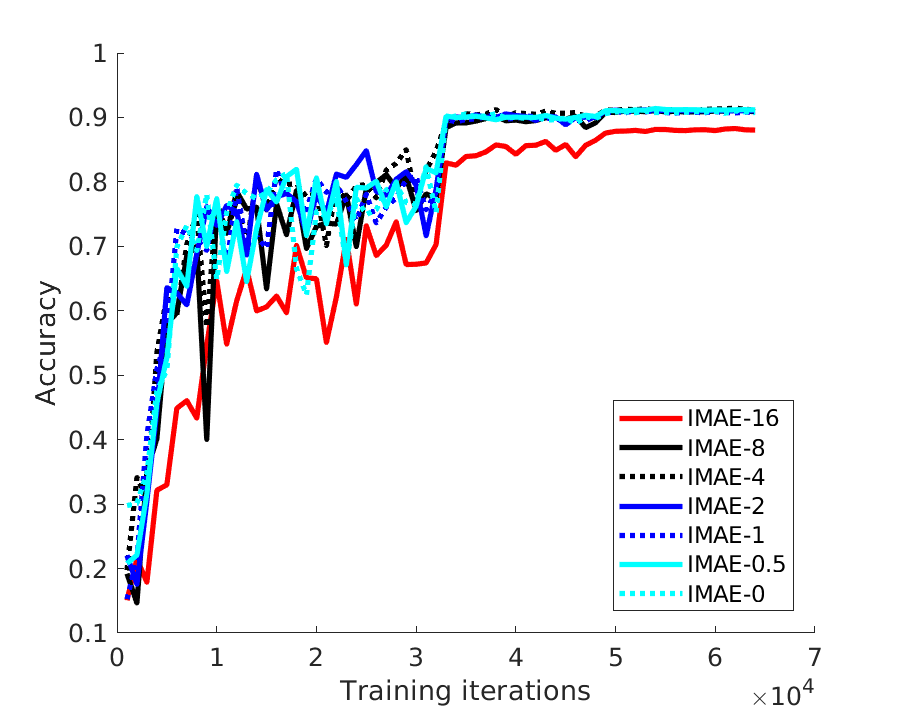}
\caption{Test accuracies of  IMAE-$T$ trained on intact training labels.}
\label{fig:exploreT_ResNet20_N00_CCE_MAE_IMAE_Ts}
\end{subfigure}
\begin{subfigure}[t!]{0.480\linewidth}
\centering
\captionsetup{width=1\linewidth}
\includegraphics[width=\linewidth]{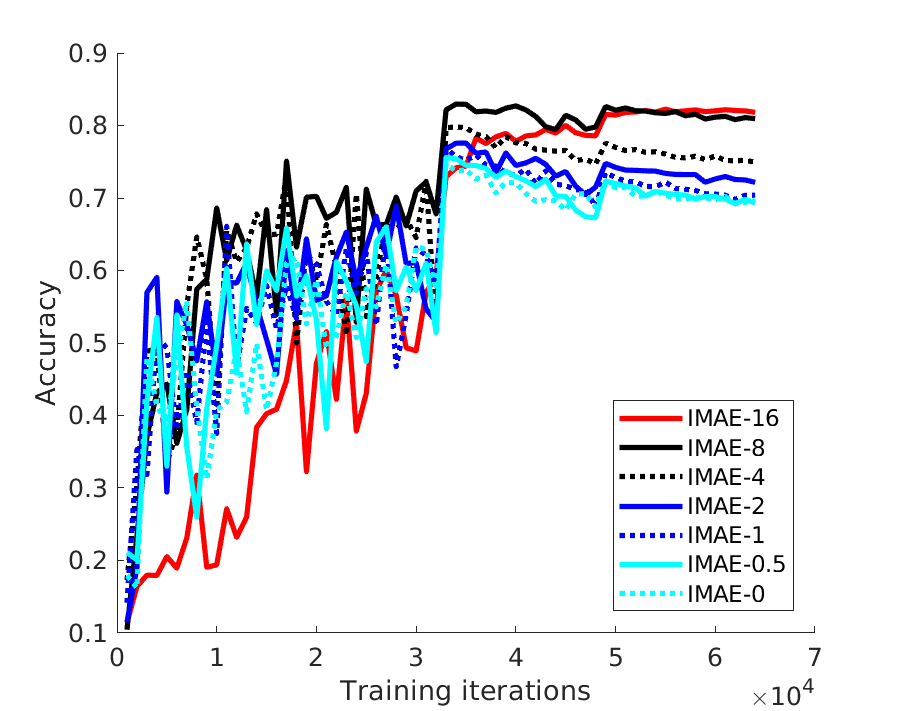}
\caption{Test accuracies of IMAE-$T$ trained on corrupted training labels.}
\label{fig:exploreT_ResNet20_N05_CCE_MAE_IMAE_Ts}
\end{subfigure}
%
%
%\vspace{-0.2cm}
\caption{The accuracy on CIFAR-10 \textit{test set} along with training iterations.
We display the results when training on intact training set and corrupted training set. 
%Test set is clean for evaluation. 
\textit{Better viewed in colour.} }
%\label{fig:ResNet2032_NoiseAll_CCE_MAE_IMAE8_corrupted_train}
%\vspace{-0.5cm}
\end{figure*}

The test results are shown and compared in Figure~\ref{fig:exploreT_ResNet20_N00_CCE_MAE_IMAE_Ts}.

{\textbf{When training labels are clean, it is unhelpful to differentiate training samples in a high degree, e.g., the performance is even lower when $T=16$. 
The final test accuracies are similar when $T$ ranges from 0 to 8.}}

%%%%%%%%%%%%%%%%%%%%%%%%%%%%%%%%%%%%%%%%%%%%%%%%%%%%%
%\begin{figure}[!h]
%	\centering
%	\includegraphics[width=0.9\linewidth]{exploreT_ResNet20_N00_CCE_MAE_IMAE_Ts}
%	\caption{
%		The test accuracies of IMAE with different $T$ on CIFAR-10 with intact training labels. 
%		\textit{Best viewed in colour.}
%	}
%	\label{fig:exploreT_ResNet20_N00_CCE_MAE_IMAE_Ts}
%	\vspace{-0.2cm}
%\end{figure}
%%%%%%%%%%%%%%%%%%%%%%%%%%%%%%%%%%%%%%%%%%%%%%%%%%%%

%%%%%%%%%%%%ResNet20_N0*_CCE_MAE_IMAE4_IMAE8%%%%%%%%%%%%%%%%%%%%%%%%%%%%
%%%%%%%%%%%ResNet20_N0*_CCE_MAE_IMAE4_IMAE8%%%%%%%%%%

%%%%%%%%%%%%ResNet20_N0*_CCE_MAE_IMAE4_IMAE8%%%%%%%%%%%%%%%%%%%%%%%%%%%%
\begin{figure*}[t!]
\centering
\begin{subfigure}[t!]{0.48\linewidth}
\centering
\captionsetup{width=1\linewidth}
\includegraphics[width=1\linewidth]{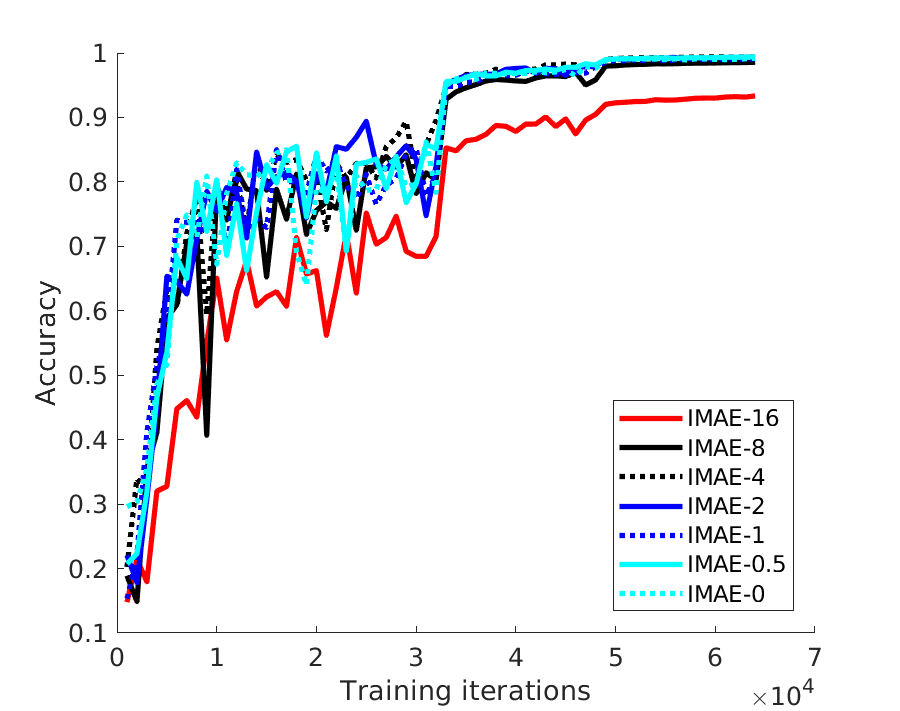}
\caption{The training accuracies of  IMAE-$T$ on intact training set.}
\label{fig:TrainAcc_exploreT_ResNet20_N00_CCE_MAE_IMAE_Ts}
\end{subfigure}
\begin{subfigure}[t!]{0.48\linewidth}
\centering
\captionsetup{width=1\linewidth}
\includegraphics[width=1\linewidth]{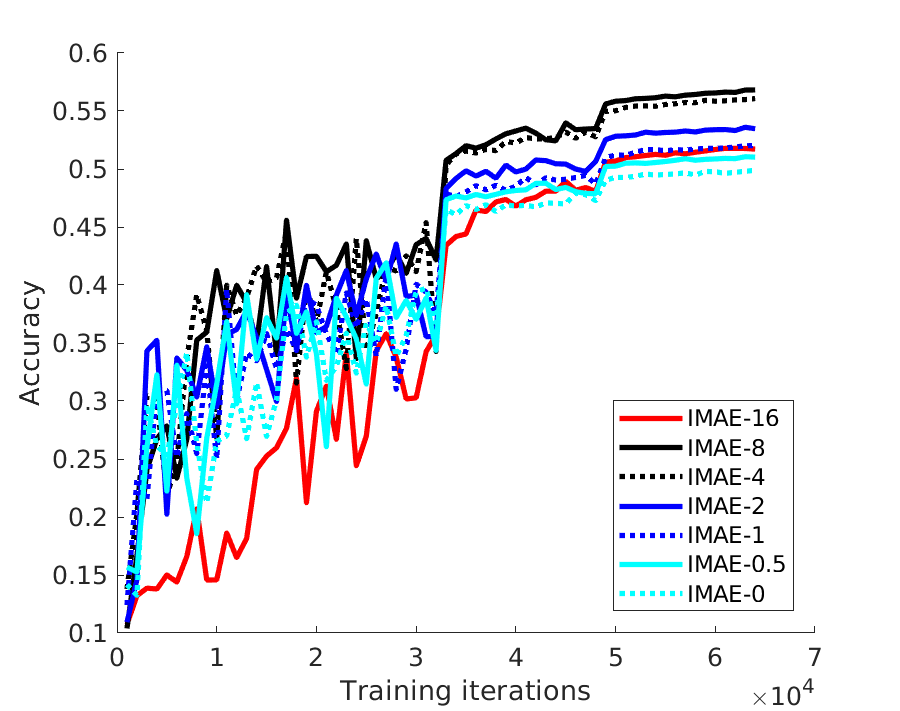}
\caption{The training accuracies of  IMAE-$T$ on corrupted training set.}
\label{fig:TrainAcc_exploreT_ResNet20_N05_CCE_MAE_IMAE_Ts}
\end{subfigure}
%
%
%\vspace{-0.2cm}
\caption{The accuracy on CIFAR-10 \textit{training sets} along with training iterations.
We show the results when training on intact training set and corrupted training set.  
\textit{Better viewed in colour.} }
\label{fig:TrainAcc_exploreT_ResNet20_N00_N05_CCE_MAE_IMAE_Ts}
%\vspace{-0.35cm}
\end{figure*}
%%%%%%%%%%%ResNet20_N0*_CCE_MAE_IMAE4_IMAE8%%%%%%%%%%

\subsection{CIFAR-10 with corrupted training labels}

The results are presented and compared in Figure~\ref{fig:exploreT_ResNet20_N05_CCE_MAE_IMAE_Ts}.
Because there exists 40\% label noise, as training goes, the test accuracy drops, which means the model overfits noisy data gradually. 

However, \textbf{we observe that higher differentiation degree (larger $T$) works better and is much less susceptible to overfitting to noisy data.} 
In Figure~\ref{fig:exploreT_ResNet20_N05_CCE_MAE_IMAE_Ts}, the final test accuracies of IMAE-16 and IMAE-8 are much higher than those of other models. 

%\vspace{-0.4cm}
\section{The Impact of $T$ on Training Accuracy}

Following the practice in the main paper, we also visualise and compare the accuracies on the training sets, which indicate how different models fit to training data as training goes, thus leading to different generalisation performance in the test phase. 
%
%All models are trained in experiments of Sec.~\ref{sec:test_accuracies}. 
We present how each model fits its corresponding training set in Figure~\ref{fig:TrainAcc_exploreT_ResNet20_N00_N05_CCE_MAE_IMAE_Ts}.

\subsection{Fitting of intact training set}

As compared in Figure~\ref{fig:TrainAcc_exploreT_ResNet20_N00_CCE_MAE_IMAE_Ts}, 
%we observe that 
\textbf{all models fit training data similarly when $T$ ranges from 0 to 8}. However, when $T=16$, the differentiation degree becomes too large as shown in Table~\ref{table:variance_IMAE_T}. {{When differentiation degree is too large, only a quite small proportion of training data can contribute.}}    
Consequently, \textbf{IMAE-16 underfits training data compared with other models.} That is why IMAE-16 has the worst test performance as shown in  Figure~\ref{fig:exploreT_ResNet20_N00_CCE_MAE_IMAE_Ts}.

\subsection{Fitting of corrupted training set}
\label{sec:fitting_of_corrupted_data}

The training accuracies of corrupted training set are displayed in Figure~\ref{fig:TrainAcc_exploreT_ResNet20_N05_CCE_MAE_IMAE_Ts}.
We have two observations:
\begin{itemize}
\item \textbf{In cases where noise rate is high, as $T$ increases, the fitting of training data first becomes better, and then becomes worse.}  
Specifically, when $T$ increases from 0 to 8, the training accuracy grows gradually, which means the fitting of training data becomes better. However, when $T=16$, the weighting variance becomes very large (Table~\ref{table:variance_IMAE_T}). 
As a result, IMAE-16's fitting of training data becomes much worse than IMAE-8's.  

\item \textbf{Fitting corrupted training data better does not mean better generalisation performance.} On the one hand, although IMAE-16 fits the training data much worse than IMAE-8 (Figure~\ref{fig:TrainAcc_exploreT_ResNet20_N05_CCE_MAE_IMAE_Ts}), IMAE-16's test accuracy is slightly better than IMAE-8's (Figure~\ref{fig:exploreT_ResNet20_N05_CCE_MAE_IMAE_Ts}). 
On the other hand, similar to IMAE-8, IMAE-4 fits its training data well (Figure~\ref{fig:TrainAcc_exploreT_ResNet20_N05_CCE_MAE_IMAE_Ts}), but IMAE-4's test performance is much worse than IMAE-8's (Figure~\ref{fig:exploreT_ResNet20_N05_CCE_MAE_IMAE_Ts}).  
%(This is different from CCE. CCE fits corrupted training data well, but its generalisation performance is terrible.)
\end{itemize}

\section{Choosing $T$ in Practice}

%%%%%%%%%%%%%%%%%%%%%%%%%%%%%%%%%%%%%%%%%%%%%%%%%%%%%
\begin{figure}[!h]
\centering
\includegraphics[width=0.65\linewidth]{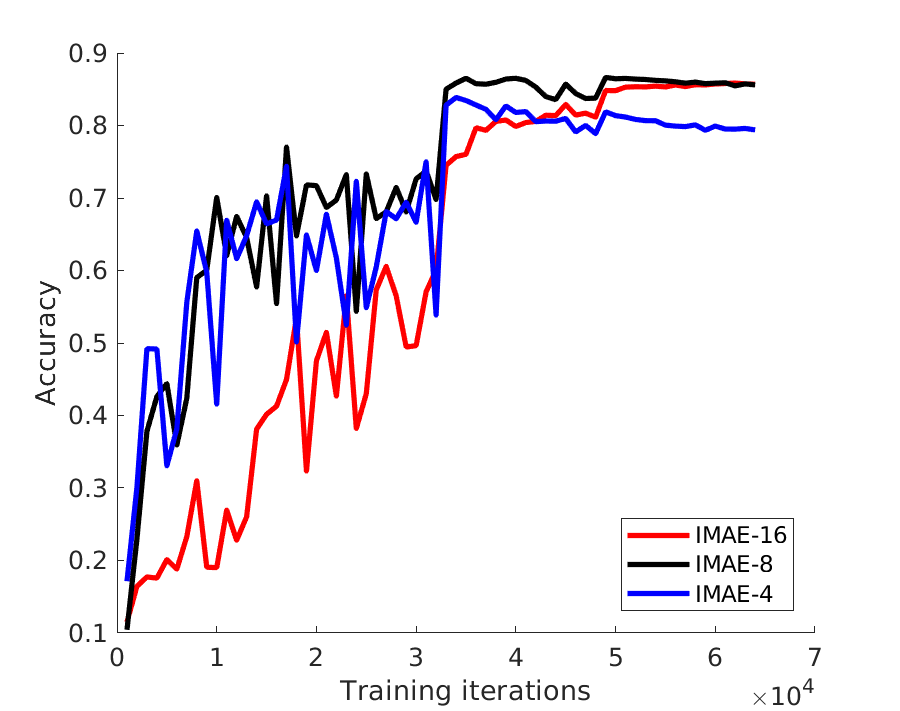}
\caption{
IMAE-16's, IMAE-8's and IMAE-4's accuracies on the clean training set when they are trained on the corrupted training set.
The overlap rate between corrupted and intact training sets is only $(1-r)= 60\%$. Therefore, \textit{we can use the original training set as a validation set}.  
\textit{Better viewed in colour.}
}
\label{fig:TrainAccIntact_exploreT_ResNet20_IMAE_Ts}
%\vspace{-0.4cm}
\end{figure}
%%%%%%%%%%%%%%%%%%%%%%%%%%%%%%%%%%%%%%%%%%%%%%%%%%%%

In summary, the training accuracy (fitting of training data) is uninformative for estimating a model's generalisation performance according to our findings in Section~\ref{sec:fitting_of_corrupted_data}. 
Therefore, it is better to optimise $T$ on a validation set in practice.

\textit{For empirical demonstration}, since the overlap rate between corrupted and intact training sets is only $(1-r)= 60\%$, \textit{{we treat the original intact training set as a validation set}}. The validation performance of IMAE-16, IMAE-8 and IMAE-4 is compared in Figure~\ref{fig:TrainAccIntact_exploreT_ResNet20_IMAE_Ts}.
We observe that IMAE-16 and IMAE-8 own similar validation performance, while IMAE-4's validation accuracy is lower. 
Furthermore, their validation performance is consistent with their test performance (Figure~\ref{fig:exploreT_ResNet20_N05_CCE_MAE_IMAE_Ts}).  
{Therefore, we conclude that it is a good practice to optimise $T$ on a validation set in different cases.}

%%%%%%%%%%%%%%%%%%%%%%%%%%%%%%%%%%%%%%%%%%%%%%%%%%%%%
\begin{figure}[!t]
\centering
\includegraphics[width=0.55\linewidth]{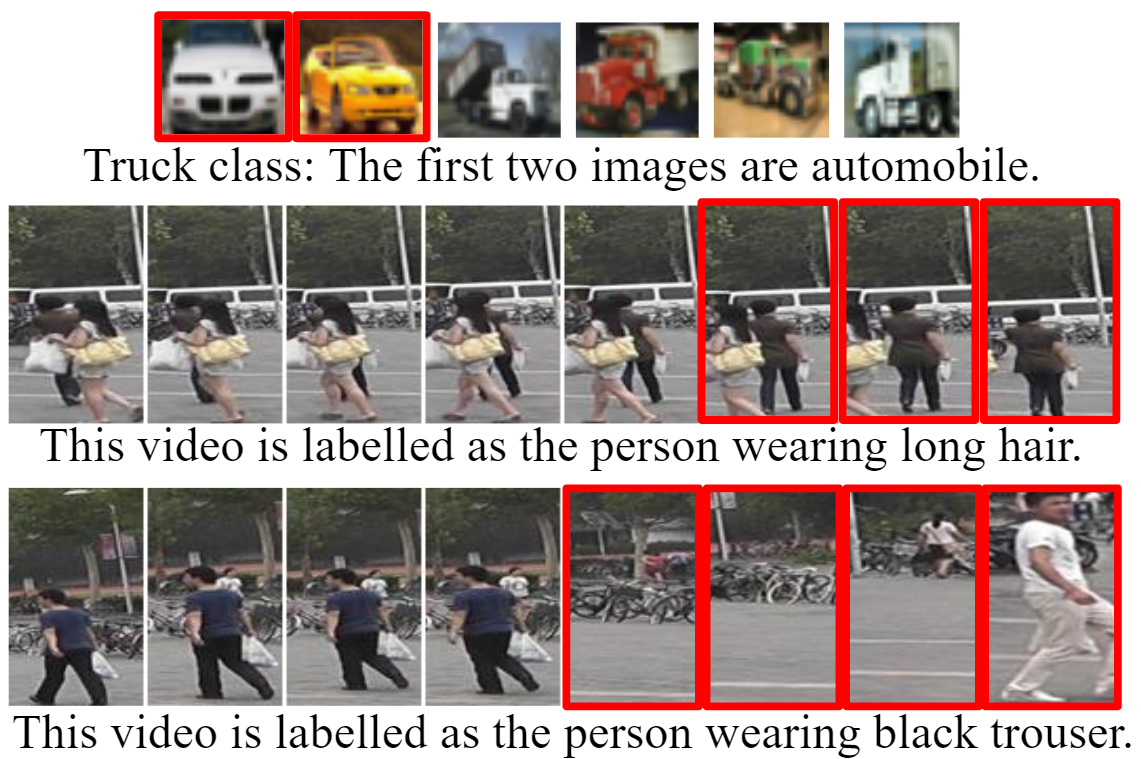}
\vspace{-0.16cm}
\caption{
Display of abnormal training examples highlighted by red boxes. 
The 1st row shows synthetic abnormal examples from corrupted CIFAR-10
~\cite{krizhevsky2009learning}.
The 2nd and 3rd rows present realistic abnormal examples from video person re-identification benchmark MARS \cite{zheng2016mars}.       
We remark three representatives: 
1) The abnormal images with no person in 3rd row contain no semantic information at all. 
2) The last abnormal image in 2nd or 3rd row may contain a person that does not belong to any person in the training set. 
3) We cannot decide the object of interest without any prior when an image contains more than one object, e.g., the 2nd and 3rd last images in 2nd row contain two persons.   	
\textit{Better viewed in colour.}
}
\label{fig:abnormal_examples}
\vspace{-0.25cm}
\end{figure}
%%%%%%%%%%%%%%%%%%%%%%%%%%%%%%%%%%%%%%%%%%%%%%%%%%%%

%%%%%%%%%%%%%%%%%%%%%%%%%%%%%%%%
\begin{table*}[!t]
%%\vspace{-0.0cm}
\caption{
The results of algorithms using different stochastic optimisers on CIFAR-10 with 40\% class-independent (symmetric) label noise.
The trained network is ResNet56 \cite{he2016deep}.   
The key hyper-parameters of all optimisers are shown. Other settings are fixed to be the same as presented in the implementation details of Section~\ref{sec:cifar10_experiments}, e.g., weight decay = 0.0001. Since Adam is an adaptive gradient method, we show several variants of it.      
}
\centering
\vspace{-6pt}
\fontsize{9pt}{9pt}\selectfont
\setlength{\tabcolsep}{6pt} % Default value: 6pt
\begin{tabular}{lcccccc}
\toprule
& \makecell{SGD \\(lr: 0.01)} & \makecell{SGD + Momentum \\ (lr: 0.01)} & \makecell{Nesterov \\(lr: 0.01)} & \makecell{Adam \\(lr: 0.01, \\delta: 0.1)} & \makecell{Adam \\(lr: 0.005, \\delta: 0.1)} & \makecell{Adam \\(lr: 0.005, \\delta: 1)} \\
\midrule
CCE & 64.3 & 60.6 & 56.4 & 42.5 & 44.5 & 50.3\\
MAE & 39.3 & 64.7 & 64.1 & 68.2 & 59.9 & 41.4\\
GCE & 68.8 & 80.5 & 79.7 & 73.2 & 70.6 & 69.3\\
IMAE & \textbf{82.0} & \textbf{83.5} & \textbf{83.7} & \textbf{75.5} & \textbf{76.3} & \textbf{78.6}\\
\bottomrule
\end{tabular}
\label{table:stochastic_optimisers}
\vspace{-0.2cm}
\end{table*}
%%%%%%%%%%%%%%%%%%%%%%%%%%%%%%%%%

\section{Video person re-identification }

\noindent
\textbf{Dataset and evaluation settings.} MARS contains 20,715 videos of 1,261 persons \cite{zheng2016mars}. 
There are 1,067,516 frames in total. 
%The image size is $128 \times 64$. 
%The videos are captured by six cameras with different views. 
%Each person is shot by at least two cameras so that cross-camera verification can be conducted. 
Because person videos are collected by tracking and detection algorithms, 
%\cite{dehghan2015gmmcp,felzenszwalb2010object}, 
abnormal examples exist as shown in Figure~\ref{fig:abnormal_examples}.
The exact noise rate is unknown.
%a lot of images contain a different person or only background \cite{wang2019deep}. 
%exactly \cite{zheng2016mars}
Following standard settings, we use 8,298 videos of 625 persons for training 
%(509,914 images in total)
and 12,180 videos of other 636 persons for testing. 
%(681,089 images in total). 
We report the cumulated matching characteristics (CMC) and mean average precision (mAP) results.

\noindent
\textbf{Implementation details.\footnote{We explore the performance of different losses in real-world applications instead of pushing the state-of-the-art results.}}  Following \cite{liu2017qan,wang2019deep}, we train GoogleNet V2. % rather than ResNet50 \cite{he2016deep}, which is less efficient. 
%follow them to
We also treat a video as an image set, which means we use only appearance information without exploiting latent temporal information. 
A video's representation is simply the average fusion of its frames' representations.
%in the tracklet.
%
\textit{We apply the same training settings for each loss}. The learning rate starts from 0.01 and is divided by 2 every 10k iterations. We stop training at 50k iterations. We choose SGD optimiser with a weight decay of 0.0005 and momentum of 0.9. The batch size is set to 180. We use standard data augmentation: a $227 \times 227$ crop is randomly sampled and flipped after resizing an original image to $256\times 256$. 
At testing, following \cite{wang2019deep,movshovitz2017no,law2017deep}, we first $L_2$ normalise videos' features and then calculate the cosine similarity between every two features.  
%The final model when training stops is used for testing.

\noindent
\textbf{Results.} We compare our method with CCE, MAE and GCE. 
%GCE is the most related algorithm and state-of-the-art.
We implement GCE with its best settings. 
%We remark that we do not corrupt any labels, thus 
%the exact noise rate is unknown. 
The results are shown in Table~\ref{table:MARS_ReID}. 
IMAE outperforms other related methods by a significant margin.
%, verifying our analysis in Sec.~\ref{sec:corrupted_cifar10}. 

%The exact noise rate is unknown. We still find that IMAE-8 performs much better than CCE, MAE, and GCE \cite{zhang2018generalized}. GCE \cite{zhang2018generalized} is a recent work proposed for increasing the noise-robustness of CCE. We reimplement its standard version (without data truncation and pruning) and set its balance parameter to $0.7$ following their practice. We observe that without data truncation and pruning, GCE's performance is between CCE and MAE, which fulfils their proposal and observation that GCE is a balance between CCE and MAE.  
%%%%%%%%%%%%%%%%%%%%%%%%%%%%%%%%
\begin{table}[!h]
%\vspace{-0.0cm}
\caption{
The retrieval results of CCE, MAE, GCE and IMAE on MARS with GoogLeNet V2 \cite{ioffe2015batch}.
}
\centering
\vspace{-0.2cm}
\setlength{\tabcolsep}{6pt} % Default value: 6pt
\begin{tabular}{lcccc}
\hline
Metric & CCE & MAE & GCE & IMAE\\
\hline
mAP (\%) & 58.1 & 12.0 & 31.6 & \textbf{70.9}\\
CMC-1 (\%) & 73.8 & 26.0 & 51.5 & \textbf{83.5}\\
\hline
\end{tabular}
\label{table:MARS_ReID}
\vspace{-0.3cm}
\end{table}
%%%%%%%%%%%%%%%%%%%%%%%%%%%%%%%%%

%%%%%%%%%%%%ResNet2056_Noise80_CCE_MAE_SMAE_dynamics%%%%%%%%%%%%%%%%%%%%%%%%%%%%
\begin{figure*}[t!]
\centering
\begin{subfigure}[t!]{0.33\linewidth}
\centering
\includegraphics[width=0.780\linewidth]{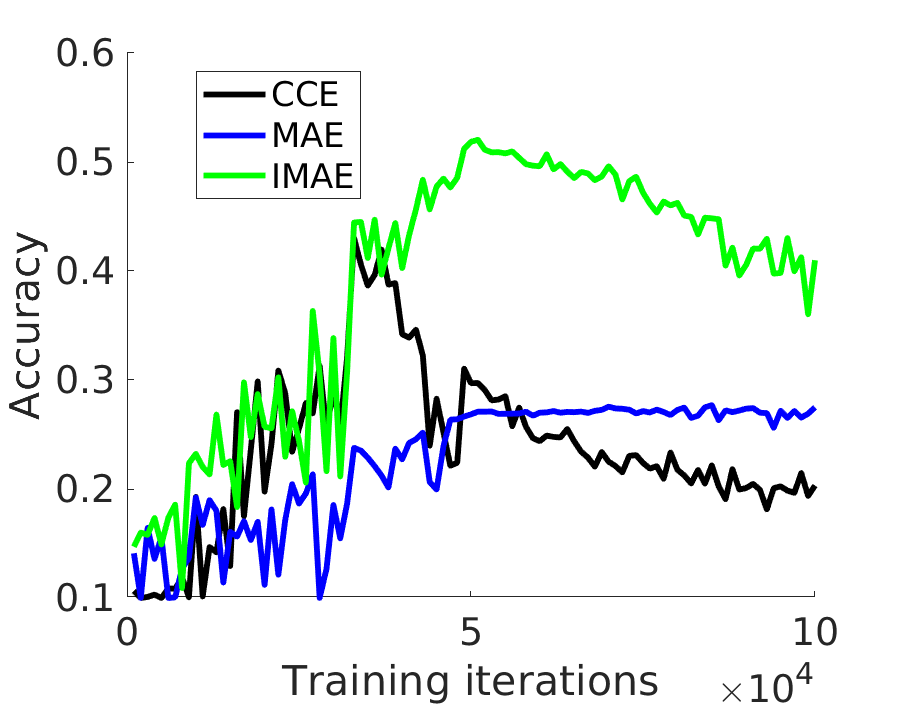}
\caption{ResNet20: Testing set (higher is better).}
\end{subfigure}%
\begin{subfigure}[t!]{0.33\linewidth}
\centering
\includegraphics[width=0.780\linewidth]{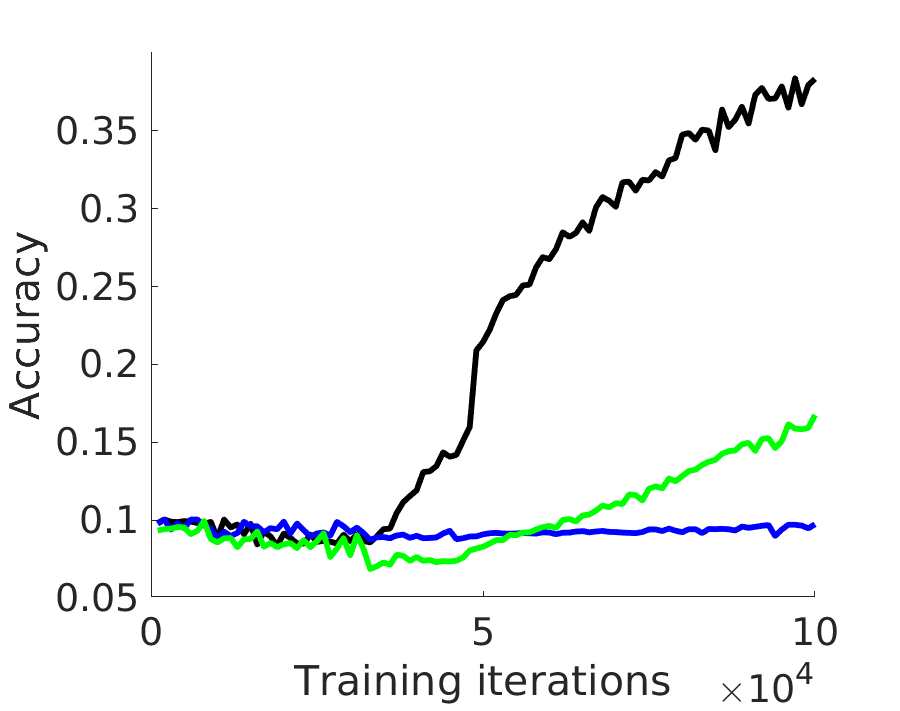}
\caption{ResNet20: Noisy subset  (lower is better).}
\end{subfigure}
\begin{subfigure}[t!]{0.33\linewidth}
\centering
\includegraphics[width=0.780\linewidth]{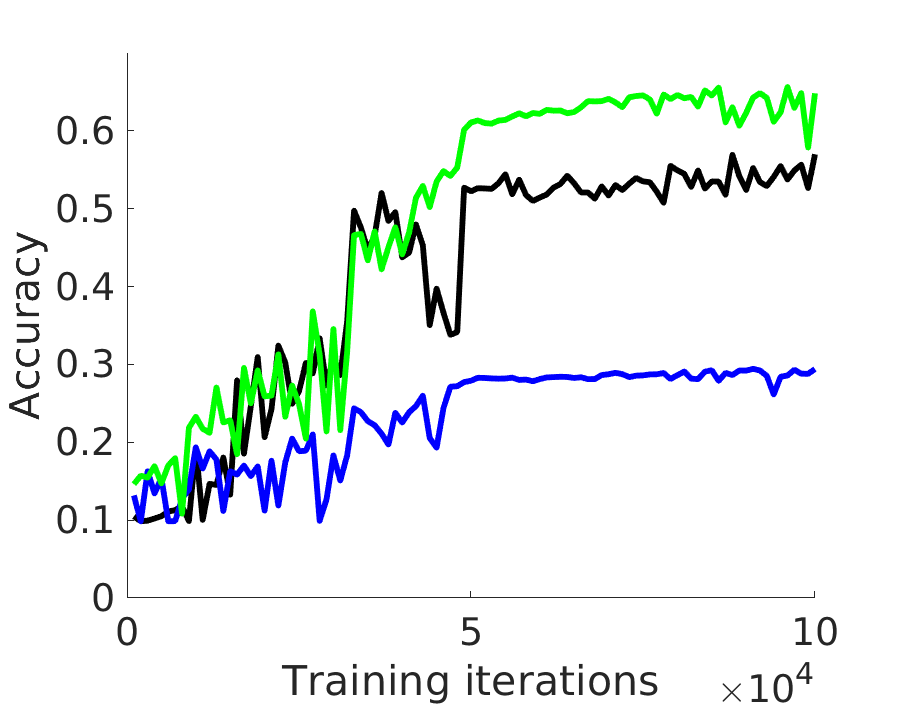}
\caption{ResNet20: Clean subset (higher is better).}
\end{subfigure}
\begin{subfigure}[t!]{0.33\linewidth}
\centering
\includegraphics[width=0.780\linewidth]{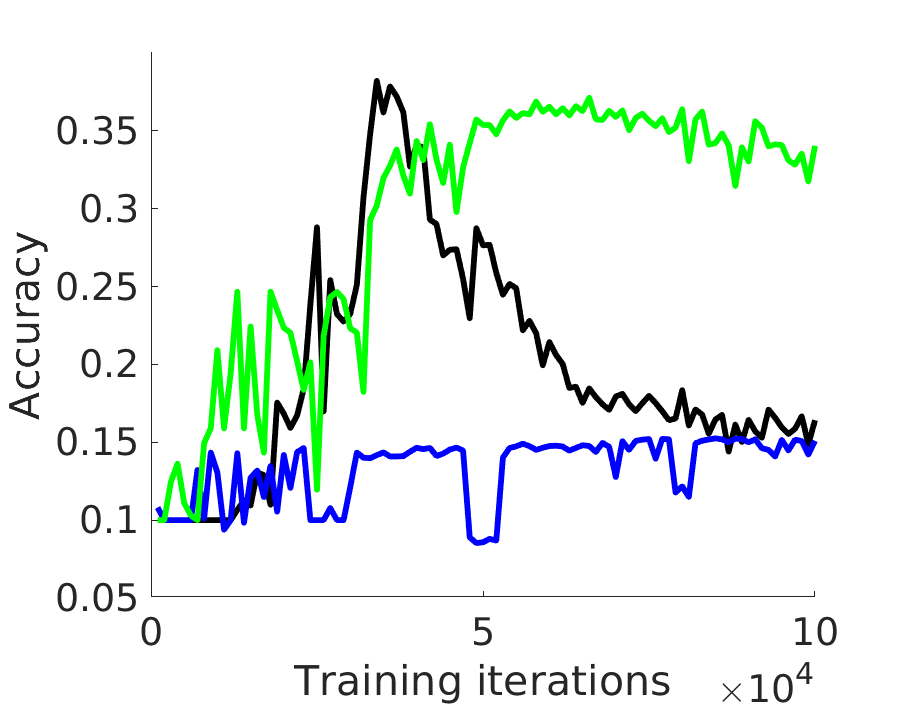}
\caption{ResNet56: Testing set (higher is better).}
\end{subfigure}%
\begin{subfigure}[t!]{0.33\linewidth}
\centering
\includegraphics[width=0.780\linewidth]{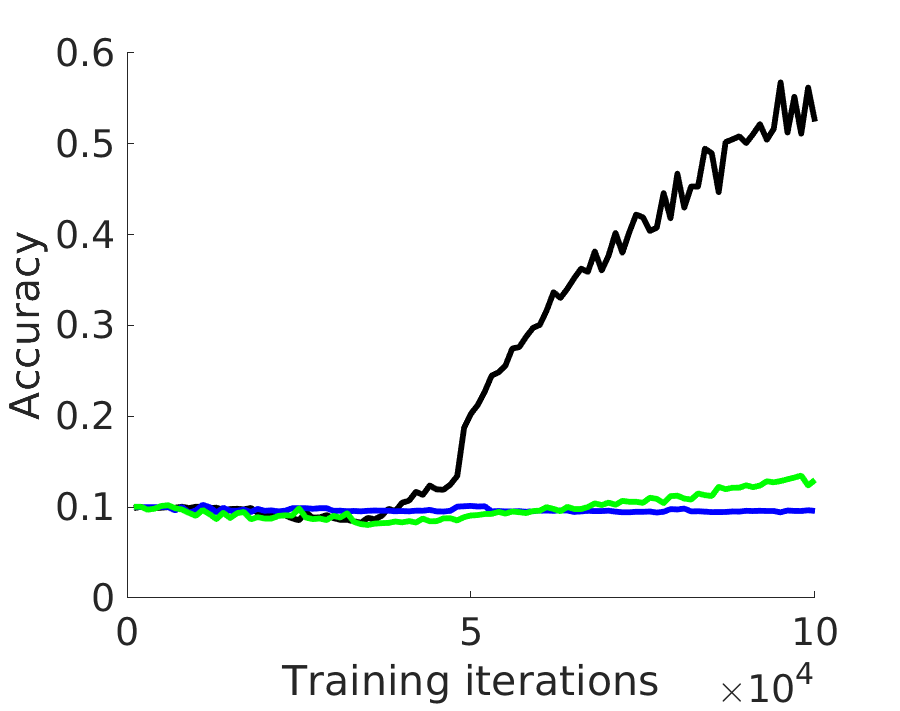}
\caption{ResNet56: Noisy subset  (lower is better).}
\end{subfigure}
\begin{subfigure}[t!]{0.33\linewidth}
\centering
\includegraphics[width=0.780\linewidth]{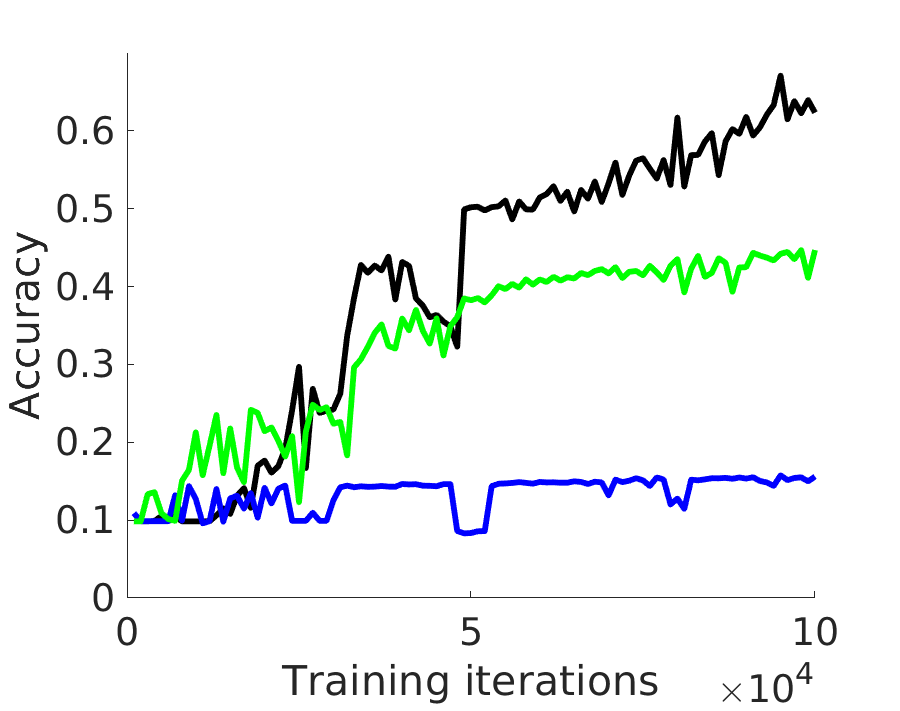}
\caption{ResNet56: Clean subset (higher is better).}
\end{subfigure}
\caption{CIFAR-10 with noise rate $r=80\%$. The accuracies on testing set, noisy subset and clean subset of training set along with training iterations.
The legend on the top left is shared by all subfigures. 
\textit{Better viewed in colour.} }
\label{fig:ResNet2056_Noise80_CCE_MAE_SMAE_dynamics}
%\vspace{-0.3cm}
\end{figure*}
%%%%%%%%%%%ResNet2056_Noise80_CCE_MAE_SMAE_dynamics%%%%%%%%%%

\section{The Results of IMAE Using Different Stochastic Optimisers}
In this section, we study the performance of IMAE when different stochastic optimisers are used. The results are presented in Table~\ref{table:stochastic_optimisers}. We observe that IMAE's results are the best consistently.

\section{Derivation of Softmax, CCE and MAE Layers}

\subsection{Derivation of softmax layer}
As the softmax layer is shared by CCE and MAE, we present the derivation of softmax layer first. 
%Given a data sample $(\mathbf{x}_i, y_i)$, the probability of $\mathbf{x}_i$ being predicted to class $y_i$ is $p(y_i|\mathbf{x}_i)$. 
First, we have
\begin{equation}
\label{eq:predict_probability}
\begin{aligned}
%\frac{1}{p(y_i|\mathbf{x}_i)} 
%&= \frac{\exp(\mathbf{z}_{iy_i}) + \sum_{j\neq y_i} \exp(\mathbf{z}_{ij}) 
%}
%{\exp(\mathbf{z}_{iy_i})}\\
%
p(y_i|\mathbf{x}_i)^{-1}
&= 1+\sum_{j\neq y_i} \exp(\mathbf{z}_{ij}-\mathbf{z}_{iy_i})
. 
\end{aligned}
\end{equation}
%Based on Eq.~(\ref{eq:predict_probability}), we will present how to calculate $\partial p(y_i|\mathbf{x}_i)/ \partial \mathbf{z}_{iy_i}$ and $\partial p(y_i|\mathbf{x}_i)/ \partial \mathbf{z}_{ij}, j\neq y_i$, respectively.

%%Based on Eq.~(\ref{eq:predict_probability}), we will calculate $\partial p(y_i|\mathbf{x}_i)/ \partial \mathbf{z}_{i}$.

%\subsubsection{$\partial p(y_i|\mathbf{x}_i)/ \partial \mathbf{z}_{iy_i}$}
\noindent
If $j = y_i$, for left and right sides of Eq.~(\ref{eq:predict_probability}), we calculate their derivatives w.r.t. $\mathbf{z}_{iy_i}$ simultaneously:
%\begin{equation}
%\label{eq:both_prob_derivation}
%\frac{-1}{p(y_i|\mathbf{x}_i)^2} 
%%{-}p(y_i|\mathbf{x}_i)^{-2} 
% \frac{\partial p(y_i|\mathbf{x}_i)}{\mathbf{z}_{iy_i}} 
%= 
%-\sum_{j\neq y_i} \exp(\mathbf{z}_{ij}-\mathbf{z}_{iy_i})
%. 
%\end{equation}
%Therefore, 
\begin{equation}
\label{eq:both_prob_derivation_final_y_i}
\begin{aligned}
&\frac{-1}{p(y_i|\mathbf{x}_i)^2} 
%{-}p(y_i|\mathbf{x}_i)^{-2} 
\frac{\partial p(y_i|\mathbf{x}_i)}{\mathbf{z}_{iy_i}} 
= 
-\sum_{j\neq y_i} \exp(\mathbf{z}_{ij}-\mathbf{z}_{iy_i})
\\
&=>
%\\
\frac{\partial p(y_i|\mathbf{x}_i)}{\mathbf{z}_{iy_i}} 
%&= p(y_i|\mathbf{x}_i)^2 \times
%\sum_{j\neq y_i} \exp(\mathbf{z}_{ij}-\mathbf{z}_{iy_i}) \\
%&= 
%p(y_i|\mathbf{x}_i)^2 \times
%(\frac{1}{p(y_i|\mathbf{x}_i)}-1) \\
= 
p(y_i|\mathbf{x}_i) 
(1-p(y_i|\mathbf{x}_i))
. 
\end{aligned}
\end{equation}

%For left and right sides of Eq.~(\ref{eq:predict_probability}), we calculate their derivatives w.r.t. $\mathbf{z}_{ij}, j\neq y_i$ simultaneously:

\noindent
If $j \neq y_i$, analogously we have:

%%\begin{equation}
%%\label{eq:both_prob_derivation}
%%%{-}p(y_i|\mathbf{x}_i)^{-2} 
%%\frac{-1}{p(y_i|\mathbf{x}_i)^2} 
%% \frac{\partial p(y_i|\mathbf{x}_i)}{\mathbf{z}_{ij}} = 
%%\exp(\mathbf{z}_{ij}-\mathbf{z}_{iy_i})
%%. 
%%\end{equation}
%Therefore, 
\begin{equation}
\label{eq:both_prob_derivation_final_j}
\begin{aligned}
&\frac{-1}{p(y_i|\mathbf{x}_i)^2} 
\frac{\partial p(y_i|\mathbf{x}_i)}{\mathbf{z}_{ij}} = 
\exp(\mathbf{z}_{ij}-\mathbf{z}_{iy_i})\\
&=>
\frac{\partial p(y_i|\mathbf{x}_i)}{\mathbf{z}_{ij}} 
%&= 
%-p(y_i|\mathbf{x}_i)^2 \times
%\exp(\mathbf{z}_{ij}-\mathbf{z}_{iy_i}) \\
%&= 
%-p(y_i|\mathbf{x}_i) \times
%\frac{\exp(\mathbf{z}_{ij})}{\sum_{j=1}^{C} \exp(\mathbf{z}_{ij})} \\
= 
-p(y_i|\mathbf{x}_i) 
p(j|\mathbf{x}_i)
. 
\end{aligned}
\end{equation}

\noindent
{In summary}, the derivation of softmax layer is:
\begin{equation}
\label{eq:both_prob_derivation_final}
\begin{aligned}
\frac{\partial p(y_i|\mathbf{x}_i)}{\partial \mathbf{z}_{ij}} 
&=
\begin{cases} 
p(y_i|\mathbf{x}_i) 
(1-p(y_i|\mathbf{x}_i))
\text{, } &j = y_i  \\
-p(y_i|\mathbf{x}_i) 
p(j|\mathbf{x}_i)  
\text{, } &j \neq y_i
\end{cases}
\end{aligned}
\end{equation}

\noindent
\subsection{Derivation of loss layer: CCE}
According to Eq.~(\ref{eq:loss_CCE}), 
we have
\begin{equation}
\label{eq:loss_CCE_x_i}
\begin{aligned}
L_{\mathrm{CCE}} (\mathbf{x}_i;f_\theta,\mathbf{W}) 
%&= 	 -\sum_{j=1}^{C} \log(p(j|\mathbf{x}_i))q(j|\mathbf{x}_i)
%\\
&= 	- \log p(y_i|\mathbf{x}_i)
.
\end{aligned}
\end{equation}
Therefore, we obtain (the parameters are omitted for brevity),   
\begin{equation}
\label{eq:derivation_CCE_x_i}
\begin{aligned}
\frac{\partial L_{\mathrm{CCE}}(\mathbf{x}_i)}{\partial p(j|\mathbf{x}_i)} 
&=
\begin{cases} 
%\frac{-1}{p(y_i|\mathbf{x}_i)} 
-p(y_i|\mathbf{x}_i)^{-1} 
\text{, } &j = y_i  \\
0       
\text{, } &j \neq y_i
\end{cases}
.
\end{aligned}
\end{equation}

%Based on Eq.~(\ref{eq:derivation_CCE_x_i}), \cite{ghosh2017robust,zhang2018generalized} concluded that CCE is sensitive to samples with corrupted labels. Their gradient could be very large as their probabilities are generally small, e.g., $1e-5$.

\noindent
\subsection{Derivation of loss layer: MAE}
According to Eq.~(\ref{eq:loss_MAE}), 
we have
\begin{equation}
\label{eq:loss_MAE_x_i}
\begin{aligned}
L_{\mathrm{MAE}} (\mathbf{x}_i;f_\theta,\mathbf{W}) 
%&= 	 -\sum_{j=1}^{C} \log(p(j|\mathbf{x}_i))q(j|\mathbf{x}_i)
%\\
&= 	2(1- (p(y_i|\mathbf{x}_i))
.
\end{aligned}
\end{equation}
Therefore, we obtain % (the parameters are omitted too),   
\begin{equation}
\label{eq:derivation_MAE_x_i}
\begin{aligned}
\frac{\partial L_{\mathrm{MAE}}(\mathbf{x}_i)}{\partial p(j|\mathbf{x}_i)} 
&=
\begin{cases} 
-2  \text{, } &j = y_i  \\
0        \text{, } &j \neq y_i
\end{cases}
.
\end{aligned}
\end{equation}

%Based on Eq.~(\ref{eq:derivation_MAE_x_i}), \cite{ghosh2017robust,zhang2018generalized} concluded that MAE treats every samples equally, thus being noise-tolerant.

\subsection{Derivatives w.r.t. $\mathbf{z}_{i}$}

%Based on Eq.~(\ref{eq:derivation_CCE_x_i}), \cite{ghosh2017robust,zhang2018generalized} concluded that CCE is sensitive to samples with corrupted labels. Their gradient could be very large as their probabilities are generally small, e.g., $1e-5$.
%Based on Eq.~(\ref{eq:derivation_MAE_x_i}), \cite{ghosh2017robust,zhang2018generalized} concluded that MAE treats every samples equally, thus being noise-tolerant.

\noindent
{$\partial L_{\mathrm{CCE}}(\mathbf{x}_i)/ \partial \mathbf{z}_{i}$}: The calculation is based on Eq.~(\ref{eq:derivation_CCE_x_i}) and Eq.~(\ref{eq:both_prob_derivation_final}).

\noindent
If $j = y_i$, we have:
\begin{equation}
\begin{aligned}
\frac{\partial L_{\mathrm{CCE}}(\mathbf{x}_i)}{\partial \mathbf{z}_{iy_i}} 
&= \sum_{j=1}^{C}  \frac{\partial L_{\mathrm{CCE}}(\mathbf{x}_i)}{\partial p(j|\mathbf{x}_i)}  \frac{\partial p(y_i|\mathbf{x}_i)}{\mathbf{z}_{iy_i}} \\
%&= \frac{-1}{p(y_i|\mathbf{x}_i)} \times p(y_i|\mathbf{x}_i) \times (1-p(y_i|\mathbf{x}_i)) \\
&=p(y_i|\mathbf{x}_i)-1
.
\end{aligned}
\end{equation}

\noindent
If $j \neq y_i$, it becomes:
\begin{equation}
\begin{aligned}
\frac{\partial L_{\mathrm{CCE}}(\mathbf{x}_i)}{\partial \mathbf{z}_{ij}} 
&= \sum_{j=1}^{C}  \frac{\partial L_{\mathrm{CCE}}(\mathbf{x}_i)}{\partial p(j|\mathbf{x}_i)}  \frac{\partial p(y_i|\mathbf{x}_i)}{\mathbf{z}_{ij}} \\
%&= \frac{-1}{p(y_i|\mathbf{x}_i)} \times (-p(y_i|\mathbf{x}_i) \times p(j|\mathbf{x}_i)) \\
&=p(j|\mathbf{x}_i)
.
\end{aligned}
\end{equation}

\noindent
{In summary}, $\partial L_{\mathrm{CCE}}(\mathbf{x}_i)/ \partial \mathbf{z}_{i}$ can be represented as:
\begin{equation}
\label{eq:summary_CCE_z}
\begin{aligned}
\frac{\partial L_{\mathrm{CCE}}(\mathbf{x}_i)}{\partial \mathbf{z}_{ij}} 
&=
\begin{cases} 
p(y_i|\mathbf{x}_i)-1  \text{, } &j = y_i  \\
p(j|\mathbf{x}_i)        \text{, } &j \neq y_i
\end{cases}
.
\end{aligned}
\end{equation}

\noindent
$\partial L_{\mathrm{MAE}}(\mathbf{x}_i)/ \partial \mathbf{z}_{i}$: 
The calculation is analogous with that of  $\partial L_{\mathrm{CCE}}(\mathbf{x}_i)/ \partial \mathbf{z}_{i}$.
According to Eq.~(\ref{eq:derivation_MAE_x_i}) and Eq.~(\ref{eq:both_prob_derivation_final}), if $j = y_i$:
\begin{equation}
\begin{aligned}
\frac{\partial L_{\mathrm{MAE}}(\mathbf{x}_i)}{\partial \mathbf{z}_{iy_i}} &= \sum_{j=1}^{C}  \frac{\partial L_{\mathrm{MAE}}(\mathbf{x}_i)}{\partial p(j|\mathbf{x}_i)} 
%\times
\frac{\partial p(y_i|\mathbf{x}_i)}{\mathbf{z}_{iy_i}} \\
&= -2 p(y_i|\mathbf{x}_i) 
%\times
(1-p(y_i|\mathbf{x}_i))
.
\end{aligned}
\end{equation}
otherwise ($j \neq y_i$):
\begin{equation}
\begin{aligned}
\frac{\partial L_{\mathrm{MAE}}(\mathbf{x}_i)}{\partial \mathbf{z}_{ij}} &= \sum_{j=1}^{C}  \frac{\partial L_{\mathrm{MAE}}(\mathbf{x}_i)}{\partial p(j|\mathbf{x}_i)} 
%\times
\frac{\partial p(y_i|\mathbf{x}_i)}{\mathbf{z}_{ij}} \\
&= 2 p(y_i|\mathbf{x}_i) 
%\times
p(j|\mathbf{x}_i)
.
\end{aligned}
\end{equation}

\noindent
{In summary}, $\partial L_{\mathrm{MAE}}(\mathbf{x}_i)/ \partial \mathbf{z}_{i}$ is:
\begin{equation}
\label{eq:summary_MAE_z}
\begin{aligned}
\frac{\partial L_{\mathrm{MAE}}(\mathbf{x}_i)}{\partial \mathbf{z}_{ij}} 
&=
\begin{cases} 
2 p(y_i|\mathbf{x}_i)
(p(y_i|\mathbf{x}_i)-1)  \text{, } &j = y_i  \\
2 p(y_i|\mathbf{x}_i)
p(j|\mathbf{x}_i)        \text{, } &j \neq y_i
\end{cases}
.
\end{aligned}
\end{equation}

%{\small
%	%\bibliographystyle{ieee_fullname}
%	\bibliography{ICML2020_DM}
%	\bibliographystyle{iclr2021_conference}
%}

%\bibliography{iclr2021_conference}
%\bibliographystyle{iclr2021_conference}
%
%
%\appendix
%\section{Appendix}
%You may include other additional sections here.

\end{document}

%% file: math_commands.tex
\usepackage{amsmath,amsfonts,bm}

\def\eqref#1{equation~\ref{#1}}
\def\1{\bm{1}}

\DeclareMathAlphabet{\mathsfit}{\encodingdefault}{\sfdefault}{m}{sl}
\SetMathAlphabet{\mathsfit}{bold}{\encodingdefault}{\sfdefault}{bx}{n}